\pdfoutput=1
\documentclass{article}

\usepackage[nonatbib,final]{nips_2017}

\usepackage[utf8]{inputenc} 
\usepackage[T1]{fontenc}    
\usepackage{hyperref}       
\usepackage{url}            
\usepackage{booktabs}       
\usepackage{amsfonts}       
\usepackage{nicefrac}       
\usepackage{microtype}      

\usepackage{graphicx}
\usepackage{multicol}
\usepackage{subfigure}

\usepackage{amsmath,amsfonts,amsthm,bm} 

\title{Bayesian Policy Gradients via Alpha Divergence Dropout Inference}
\author{
  Peter Henderson* \hspace{1mm}
  Thang Doan* \hspace{1mm}
  Riashat Islam \hspace{1mm}
  {\bf David Meger} \\
 * Equal contributors \\
   McGill University \hspace{2mm}\\
  \texttt{peter.henderson@mail.mcgill.ca} \hspace{2mm}
  \texttt{thang.doan@mail.mcgill.ca} \\
  \texttt{riashat.islam@mail.mcgill.ca} \hspace{2mm}
  \texttt{david.meger@mcgill.ca} \\
}

\begin{document}

\maketitle

\begin{abstract}

Policy gradient methods have had great success in solving continuous control tasks, yet the stochastic nature of such problems makes deterministic value estimation difficult. We propose an approach which instead estimates a distribution by fitting the value function with a Bayesian Neural Network. We optimize an $\alpha$-divergence objective with Bayesian dropout approximation to learn and estimate this distribution. We show that using the Monte Carlo posterior mean of the Bayesian value function distribution, rather than a deterministic network, improves stability and performance of policy gradient methods in continuous control MuJoCo simulations.


\end{abstract}

\section{Introduction}
\pdfoutput=1

Reinforcement learning (RL) has recently had great success in solving complex tasks with continuous control~\cite{schulman2015trust,schulman2017proximal,DDPG}. However, as these methods can be high variance and often deal with unstable environments, distributional perspectives of function approximations in RL have begun to gain popularity~\cite{bellemare2017distributional,BootstrappedDQN}. Currently, such methods in RL typically use many approximators (usually with shared hidden layers) to fit a distribution. However, in other fields, recent advances in Bayesian inference using deep neural networks have had notable success in providing predictive uncertainty estimates~\cite{gal2016dropout, blundell, PBP, Krueger}. Particularly, it has been shown that dropout can be used as a variational Bayesian approximation~\cite{gal2016dropout,li2017dropout}. Such dropout approximations of uncertainty estimates have demonstrated improved performance from domains such as simple classification tasks to active learning~\cite{gal2016improving}.

In this work, we develop an approach to using Bayesian neural networks (BNNs) as value function approximators in model-free policy gradient methods for continuous control tasks. We use a technique for dropout inference in BNNs using $\alpha$-divergences~\cite{li2017dropout} that provides accurate approximation of uncertainty and a Monte Carlo objective which can simulate a Gaussian distribution. We show that by using the posterior mean of this uncertainty distribution during value estimation, we achieve more stable learning and significantly better results versus a standard deterministic function approximator. We demonstrate the significance of using Monte Carlo dropout
approximation across a range of policy gradient methods including Trust Region Policy Optimization (TRPO)~\cite{schulman2015trust} and Proximal Policy Optimization (PPO)~\cite{schulman2017proximal}, and Deep Deterministic Policy Gradients (DDPG), on a variety of benchmark continuous control tasks from OpenAI Gym~\cite{brockman2016openai} using the MuJoCo simulator~\cite{mujoco}).

\section{Background and Related Work}
\pdfoutput=1

\subsection{Policy Gradient Methods for Continuous Control}

Policy gradient (PG) methods \cite{sutton2000policy} are a form of reinforcement learning which can utilize stochastic gradient descent to optimize a parameterized policy $\pi_\theta$. Such methods optimize the discounted return: $\rho (\theta, s_0) = \mathbb{E}_{\pi_\theta}\left[ \sum_{t=0}^\infty \gamma^t r(s_t) | s_0 \right]$, such that the overall policy gradient theorem results in: $\frac{\delta \rho (\theta, s_0)}{\delta \theta} = \sum_s \mu_{\pi_\theta} (s | s_0) \sum_a \frac{\delta \pi_\theta (a | s)}{\delta \theta} Q_{\pi_\theta} (s,a)$, where $\mu_{\pi_\theta} (s | s_0) = \sum_{t=0}^\infty \gamma^t P(s_t = s | s_0)$ \cite{sutton2000policy}. Trust Region Policy Optimization (TRPO) \cite{schulman2015trust} and Proximal Policy Optimization (PPO) \cite{schulman2017proximal} constrain these updates via trust regions. Furthermore, these methods leverage advantage estimation to reduce variance in updates. Generally, for these trust region methods, updates are as follows: $\max_\theta \mathbb{E}_{t} \left[ \frac{\pi_\theta (a_t|s_t)}{\pi_{\theta_{old}}(a_t|s_t)} A_t (s_t, a_t) \right]$ subject to $\mathbb{E}_t \left[\text{KL} \left[ \pi_{\theta_{old}} ( \cdot | s_t), \pi_\theta (\cdot | s_t) \right] \right] \le \delta$. $A_t (s_t, a_t)$ is the advantage function. TRPO uses constrained conjugate gradient descent to solve the constrained optimization problem. PPO transforms the constraint into a penalty or clipping objective, depending on the implementation. Here, we look exclusively at the clipping objective. Finally, DDPG~\cite{DDPG} leverages the policy gradient theorem in an actor-critic format such that a deterministic policy can be used and off-policy sampling is utilized. This yields faster training, but often at the cost of higher variance and instability~\cite{rlmatters}.

\subsection{Dropout as Bayesian Inference}

Several works demonstrate different ways to approximate an uncertainty distribution (as with a Bayesian Neural Network) via simple dropout~\cite{gal2016dropout,li2017dropout}. In~\cite{gal2016dropout}, it is shown that fitting a dropout objective (and running dropout at test time) can approximate a variational Bayesian approximation (and thus an uncertainty estimate). This work was further expanded in~\cite{li2017dropout}, where it was shown that by fitting a Monte Carlo variational inference objective utilizing the $\alpha$-divergence, an improved uncertainty estimate can be achieved. Such dropout Bayesian approximation has shown success in \emph{model-based} RL already. When fitting a dynamics model with a dropout uncertainty estimate, typical Gaussian processes can be replaced with deep function approximations which tend to improve model -- and in turn policy -- performance~\cite{gal2016improving,gal2017concrete}.

\subsection{Distributional Perspectives in RL}
Several recent works have investigated various methods for modeling distributions in the context of reinforcement learning. In~\cite{BayesianAC}, a Bayesian framework was applied to the actor-critic architecture by fitting a Gaussian Process (GP) for the critic which allowed for a closed-form derivation of update rules. In~\cite{gal2016dropout}, dropout uncertainty estimates of a $Q$-value function were used for Thompson sampling and optimistic exploration. \cite{bellemare2017distributional} uses value function update rules according to a distributional perspectives. Similarly, in \cite{BootstrappedDQN}, the authors use $k$-heads on the $Q$-value to model a distribution rather than the dropout approach of~\cite{gal2016dropout}.
Here, we build on these works and successfully provide a simple replacement for fitting a BNN value function using a dropout $\alpha$-divergence objective, that, without significant modification to the core policy gradient approach, can generally improve performance in continuous control tasks.

\section{Bayesian Value Functions in Policy Gradient Reinforcement Learning}

\pdfoutput=1

\subsection{Variational Inference for Value Functions}

As in~\cite{gal2016dropout,li2017dropout}, we estimate uncertainty through approximate dropout variational inference. This involves using Monte Carlo dropout sampling to approximate a posterior distribution $q_\theta(\omega)$ where $\omega$ corresponds to a set of random weight matrices in a neural network $\omega=(W_{i})_{i=1}^{L}$ where $L$ is the number of hidden layers. This distribution is typically fit by minimizing the KL divergence of the estimated (dropout) distribution with the true distribution. However, in~\cite{li2017dropout}, it is found that rather than minimizing the KL divergence in variational inference, a better uncertainty estimate can found by using the generalized $\alpha$-divergence distance metric. Hence, following this process, we minimize an $\alpha$ energy, which with MC Dropout sampling becomes~\cite{li2017dropout} :




\begin{equation}
\mathcal{L}_{\alpha}^{MC}=-\frac{1}{\alpha}{\displaystyle \sum_{n=1}^{N}}log\,sum\,exp[-\frac{\alpha\tau}{2}\parallel y_{n}-V^{\hat{\omega_{k}}}(x_{n})\parallel_{2}^{2}]+\frac{ND}{2}log\tau{\displaystyle +p_{i}\,\sum_{i=1}^{K}}\parallel M_{i}\parallel_{2}^{2}
\label{eq:alpha}
\end{equation}

where $\tau$ is the precision of the model, $\hat{\omega}_{k}$ are
the sampled dropout weights, $\{V^{\hat{\omega_{k}}}(x_{n})\}_{k=1}^{K}$
are a set of $K$ stochastic forward passes through the neural network, and $M_{i}$ are the neural network weights without dropout. We will refer to the BNN fit with this objective as an $\alpha$-BNN.

\pdfoutput=1

\subsection{TRPO and PPO}

We first examine policy gradient methods which use advantage estimation, where the value function is used as a baseline. We focus on two such methods: PPO \cite{schulman2017proximal} and TRPO \cite{schulman2015trust}. To learn a policy, these optimize $\mathbb{E}_{s\sim \rho_{old}} \left[\frac{\pi_{\theta}(a_{t}|s_{t})}{\pi_{\theta_{old}}(a_{t}|s_{t})}\hat{A_{t}}^{GAE(\gamma,\lambda)}\right]$ where  $\hat{A_{t}}^{GAE(\gamma,\lambda)}$ is the generalized advantage estimate~\cite{GAE} given as:

\begin{equation}
\hat{A_{t}}^{GAE(\gamma,\lambda)}={\displaystyle \sum_{l=0}^{\infty}(\gamma\lambda)^{l}\delta_{t+l}}
\end{equation}

Where $\delta_{t+l}=-V(s_{t})+r_{t}+\gamma r_{t+1}+..+\gamma^{k}V(s_{t+k})$. Although the use of the advantage function reduces prediction variance, this step is by definition an uncertain approximation made using limited data. As uncertain estimates accumulate through Bellman updates over time, this has been shown to yield over estimation \cite{thrun1993bias}. Using an $\alpha$-BNN value function, we can account for this to some extent by modeling the uncertainty distribution.
Hence, we can define a Bayesian GAE function as:

\begin{equation}
\mathbb{E}[\hat{A_{t}}^{GAE(\gamma,\lambda)}|\mathcal{D}]=\mathbb{E}_{\omega\sim q(\omega)}[\hat{A_{t}}^{GAE(\gamma,\lambda)}(s,a;\omega)]={\displaystyle \sum_{l=0}^{\infty}(\gamma\lambda)^{l}\mathbb{E}_{\omega\sim q(\omega)}[\delta_{t+l}}(\omega)]
\end{equation}

where ${\displaystyle \mathbb{E}_{\omega\sim q(\omega)}[\delta_{t+l}}(\omega)]=\int_{\omega}q(\omega)[-V(s_{t};\omega)+r_{t}+\gamma r_{t+1}+..+\gamma^{k}V(s_{t+k};\omega)]d\omega$. Here, the value functions can be approximated by performing $K$ stochastic forward passes and taking the average of the posterior distribution, similarly to~\cite{li2017dropout}. This posterior mean models the uncertainty the agent has about the value of a given state. To fit the $\alpha$-BNN value function ($V^\alpha(s)$), we simply use the objective from Equation~\ref{eq:alpha}, where the target $y_i = r_{t}+\gamma r_{t+1}+..+\gamma^{k}V(s_{t+k})$, as in~\cite{schulman2017proximal}.



\pdfoutput=1

\subsection{DDPG}

Next, we investigate using an $\alpha$-BNN Q-value function in the off-policy actor-critic method, DDPG~\cite{DDPG}. Again, we use the same loss as in Equation~\ref{eq:alpha} to fit an uncertainty estimate of the action-value function. As with $V^\alpha(s)$, the $\alpha$-BNN Q function ($Q^\alpha(s,a)$) is fit such that the target corresponds to $y_i = r_t + \gamma Q'(s_{t+1},\mu'(s_{i+1} | \theta^{\mu'}) | \theta^{Q'}$. Where $\mu'$ is the target policy and $Q'$ in this case corresponds to a target network updated with soft updates as in~\cite{DDPG}.

With this Bayesian approch the DDPG update is given by (see appendix for proof):



\begin{eqnarray}
\nabla_{\theta}\mathbb{E}[J(\mu_{\theta})|\mathcal{D}] &=& \mathbb{E}_{s\sim\rho^{\mu},\omega\sim q(\omega)}[\nabla_{\theta}\mu_{\theta}(s)\nabla_{a}Q^{\pi}(s,a;\omega)]|_{a=\mu_{\theta}(s)}\\
&=& [\int_{S}\rho^{\mu}(s)\nabla_{\theta}\mu_{\theta}(s)\int_{\omega}q(\omega)Q^{\pi}(s,a;\omega)d\omega ds]|_{a=\mu_{\theta}(s)}
\end{eqnarray}

and the gradient is approximated using the Monte Carlo samples, where where $\hat{\omega}_{j,k}$ is the sampled weight ( single forward
pass) of the layer $j$, $K$ is the number of samples and $M$ the batch size.

\begin{equation}
\nabla_{\theta}\mathbb{E}[J(\mu_{\theta})|\mathcal{D}]\simeq\frac{1}{M}{\displaystyle \sum_{i=1}^{M}\nabla_{\theta}\mu_{\theta}(s_{i})[\frac{1}{K}{\displaystyle \sum_{k=1}^{K}}\nabla_{a}Q^{\pi}(s_{i},a_{i};\hat{\omega}_{1,k},..\hat{\omega}_{L,k})\mid_{a_{i}=\mu(s_{i})}}]
\end{equation}

\section{Experiments and Results}
\pdfoutput=1

To evaluate our use of $\alpha$-BNN value functions, we use MuJoCo locomotion tasks~\cite{mujoco} provided by OpenAI Gym~\cite{brockman2016openai}. We use hyperparameters based on analysis in \cite{rlmatters} (see Appendix for detailed setup). For each algorithm we run 10 trials with 10 different random seeds\footnote{Due to the long runtime of $Q^\alpha$ DDPG, we limit our experiments to HalfCheetah-v1 and 5 random seeds.}. We compare our $\alpha$-BNN value function methods against the baseline implementation as of PPO and TRPO provided with~\cite{schulman2017proximal} and DDPG provided with~\cite{plappert2017parameter}. We additionally compare against a modified version of those algorithms which only uses \textit{L2} regularization on the value function. We do so to determine the effect of the regularization term from Equation~\ref{eq:alpha}.

\begin{figure}[!htbp]
    \centering
    \includegraphics[width=.32\textwidth]{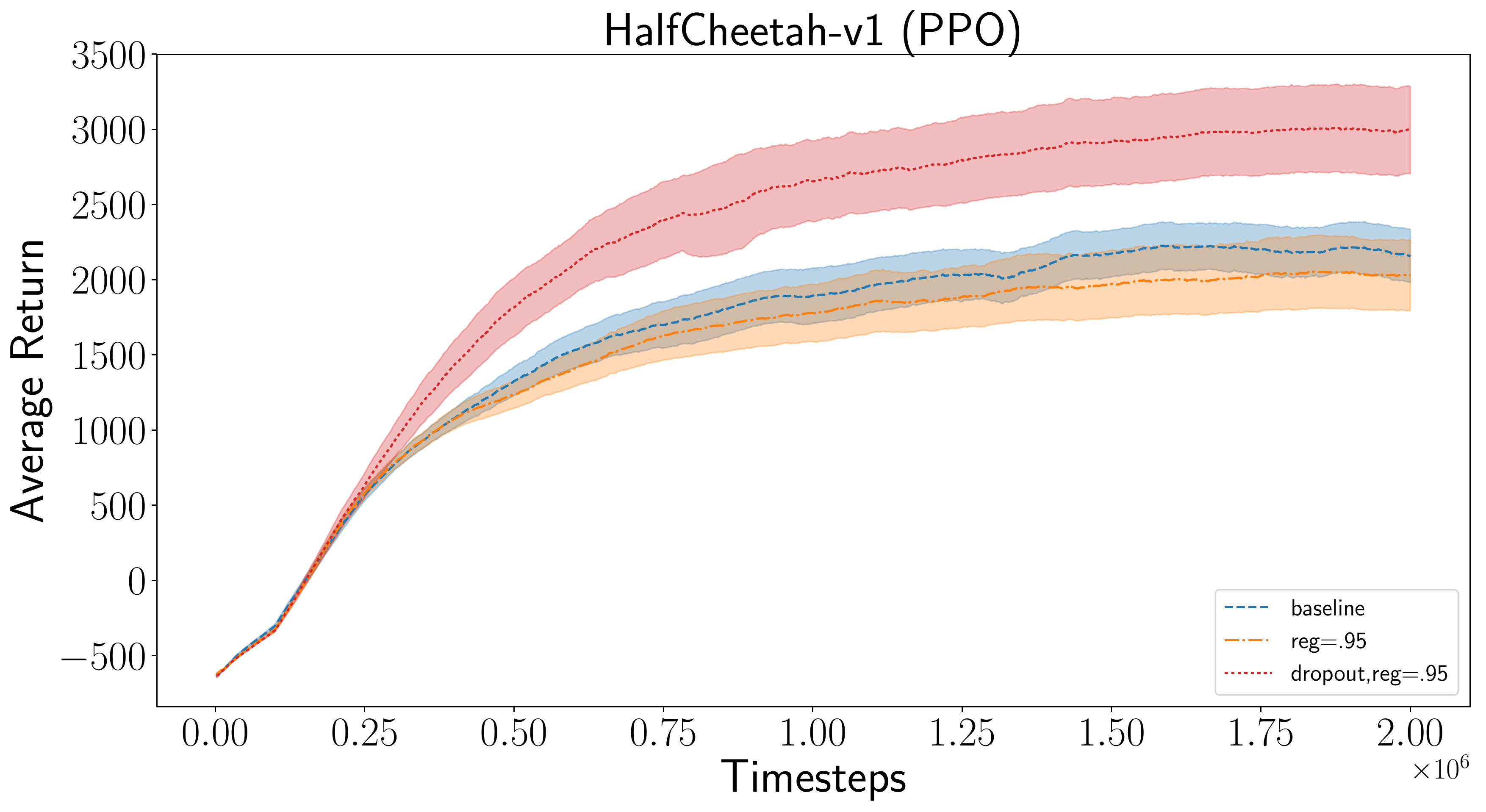}
    \includegraphics[width=.33\textwidth]{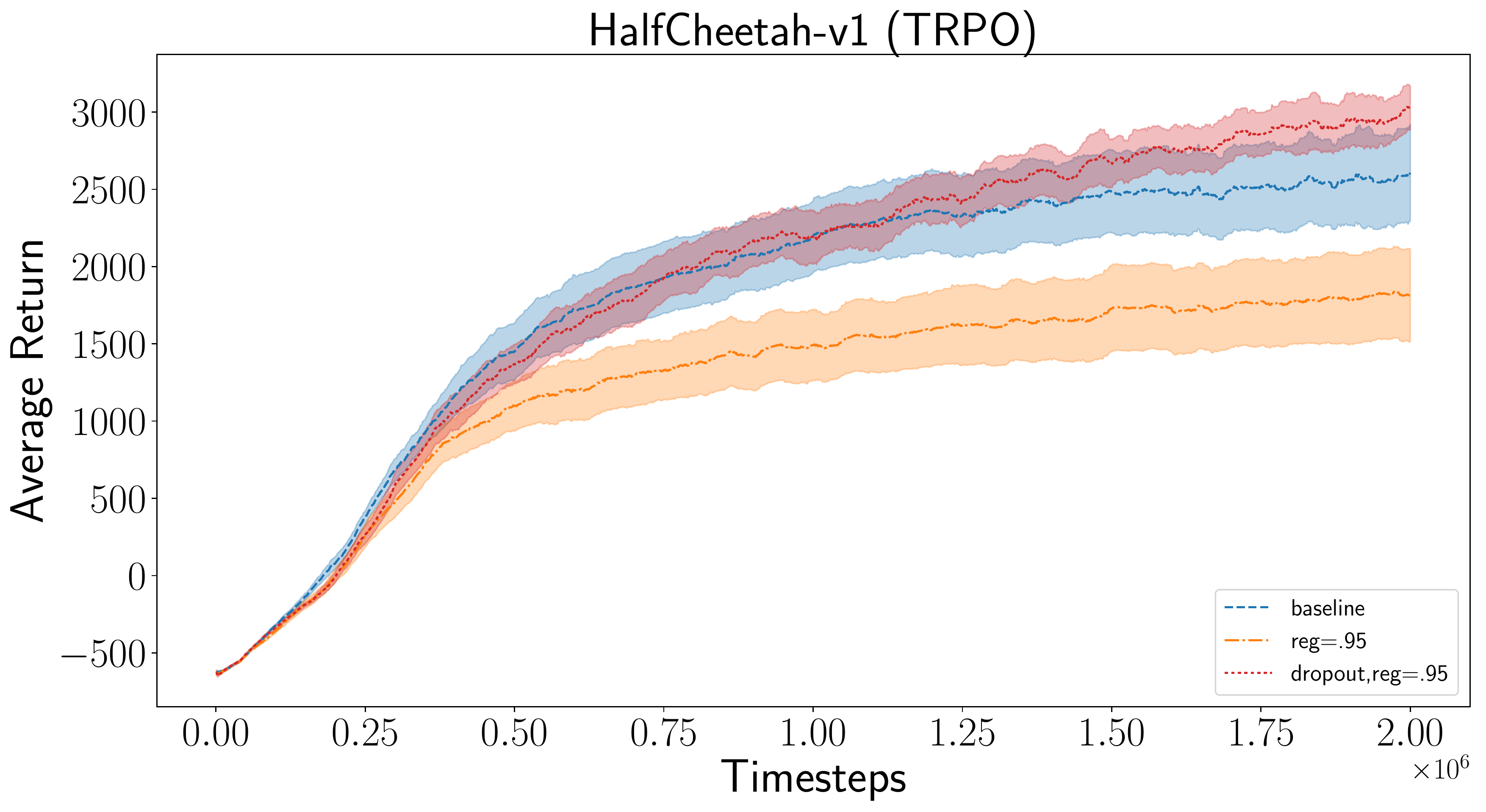}
    \includegraphics[width=.33\textwidth]{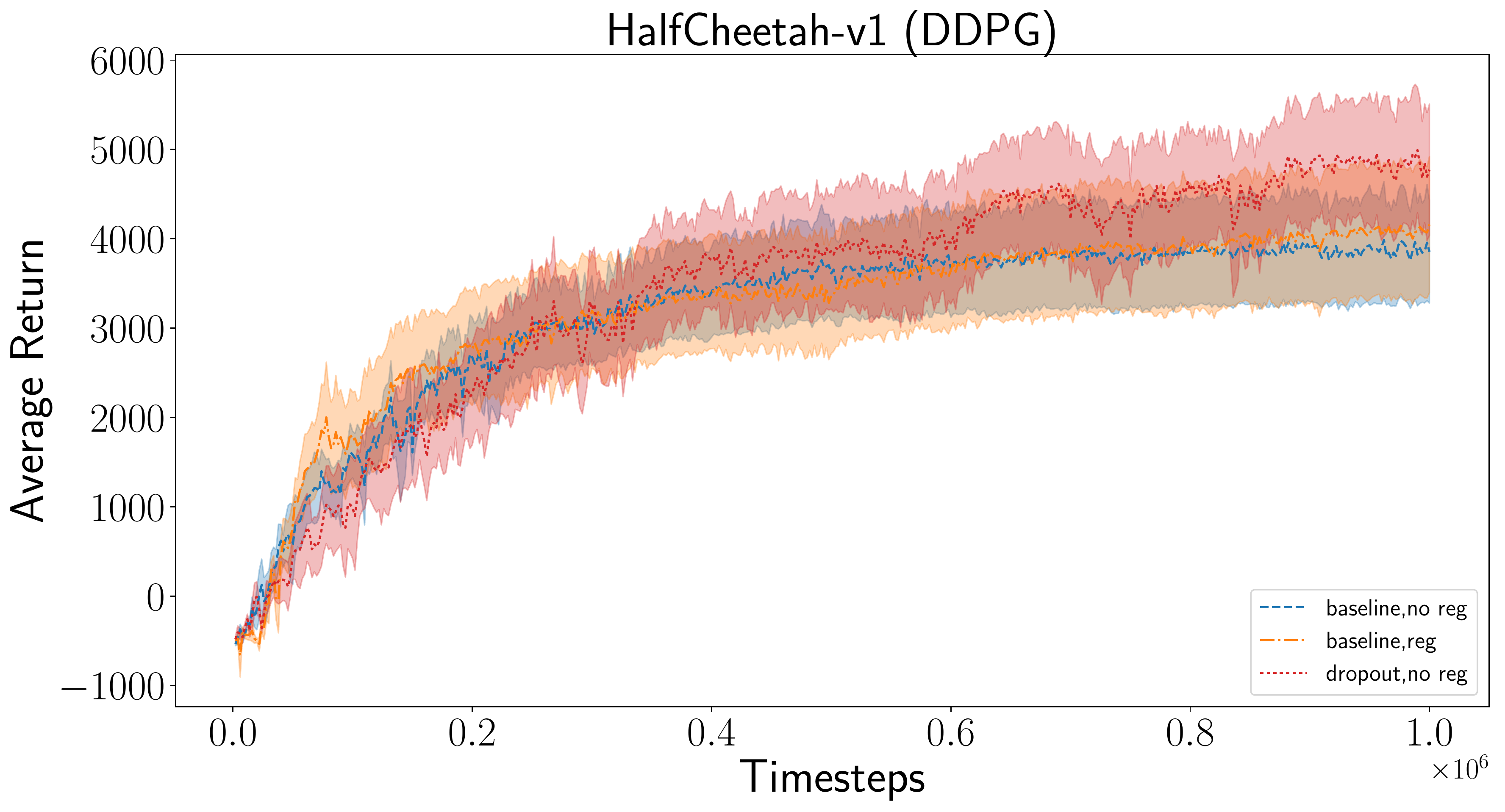}
    \caption{Performance on HalfCheetah-v1 across all three algorithms.}
    \label{fig:hc}
\end{figure}

Table~\ref{tab:ppo} summarizes our results for using the $\alpha$-BNN value function in PPO and TRPO respectively. Experimental results with $\alpha$-BNN value functions are demonstrated across Hopper-v1, HalfCheetah-v1, and Walker2d-v1 environments. Figure~\ref{fig:hc} shows learning curves for the HalfCheetah-v1 environment on all algorithms. Detailed results can also found in the Appendix. We present here optimal hyperparameters specific to the $\alpha$-divergence objective \emph{only} found via a grid search with ablation analysis on effects of individual parameters in the appendix. Shared hyperparameters between the baseline are all held constant at the default levels as provided in the OpenAI baselines implementation (this includes learning rate, maximum KL, clip params, network architecture, etc.). The only difference through all experiments from the original baselines implementation is that we use \emph{relu} activations, instead of \emph{tanh}.

\begin{table}[!htbp]
    \centering
    {\small \begin{tabular}{|c|c|c|c|}
    \hline
        &$V(s)$ & $V_{reg}(s)$ & $V_{reg}^\alpha(s)$  \\

         \hline
             \multicolumn{4}{|c|}{PPO}\\
    \hline
    Hopper-v1 & $2342 \pm 124$ & $2228 \pm 220$ & \textbf{2608 $\pm$ 79}  \\
    HalfCheetah-v1 & $2155 \pm 177$ & $2030 \pm 234$ & \textbf{2790 $\pm$ 284}$^*$  \\
     Walker2d-v1 & $3165 \pm 344$ & $3219 \pm 137$ & \textbf{3593 $\pm$ 228}$^*$  \\
         \hline
        \multicolumn{4}{|c|}{TRPO}\\
             \hline
Hopper-v1 & $1989 \pm 213$ & \textbf{2400 $\pm$ 137}& $2237 \pm 193$  \\
HalfCheetah-v1 & $2605 \pm 313$ & $1814 \pm 300$ & \textbf{3026 $\pm$ 144}$^*$  \\
Walker2d-v1 & $2974 \pm 171$ & $2328 \pm 172$ & \textbf{3406 $\pm$ 134}$^*$  \\
         \hline
                      \multicolumn{4}{|c|}{DDPG}\\
             \hline
        HalfCheetah-v1&$4159 \pm 762$&$3854 \pm 575$& \textbf{4772 $\pm$ 736}\\
         \hline
    \end{tabular}}
    \vspace{.5em}
    \caption{Final average return $\pm$ standard error across all random seeds (10 for TRPO/PPO, 5 for DDPG). Note, we do not use regularization for DDPG ($reg=0$). Significant improvements over both L2 regularization and the baseline according to 2-sample $t$-test method~\cite{rlmatters} ($p < 0.05$) marked with an asterisk ($*$). }
    \label{tab:ppo}
\end{table}




\section{Discussion}
\pdfoutput=1

Our results show significant improvements in many cases simply by fitting an $\alpha$-BNN value function and using the posterior mean during policy updates. In particular, we find large improvements in Proximal Policy Optimization from using the posterior mean. We suspect that these improvements are partially due to the added exploration, aiding the on-policy and adaptive nature of PPO, during early stages of learning where the uncertainty distribution is large. This is backed by ablation analysis (see Appendix) where we find that PPO in the Half-Cheetah environment sees larger improvements from a higher dropout rate (presumably yielding slightly more exploratory value estimates). In the overall ablation analysis (see Figure~\ref{fig:miniabl} for an example, with all results in Appendix), we found that the number of Monte Carlo samples, the dropout rate, and the $\tau$ factor generally yielded the largest difference in results. The $\tau$ parameter can be thought of as the trade-off between optimization of the objective loss and regularization. As such, we can see variations in performance due to emphasis on these different components, such that too small a $\tau$ can hurt performance by over-emphasizing regularization.

\begin{figure}[!htbp]
    \centering
    \includegraphics[width=.49\textwidth]{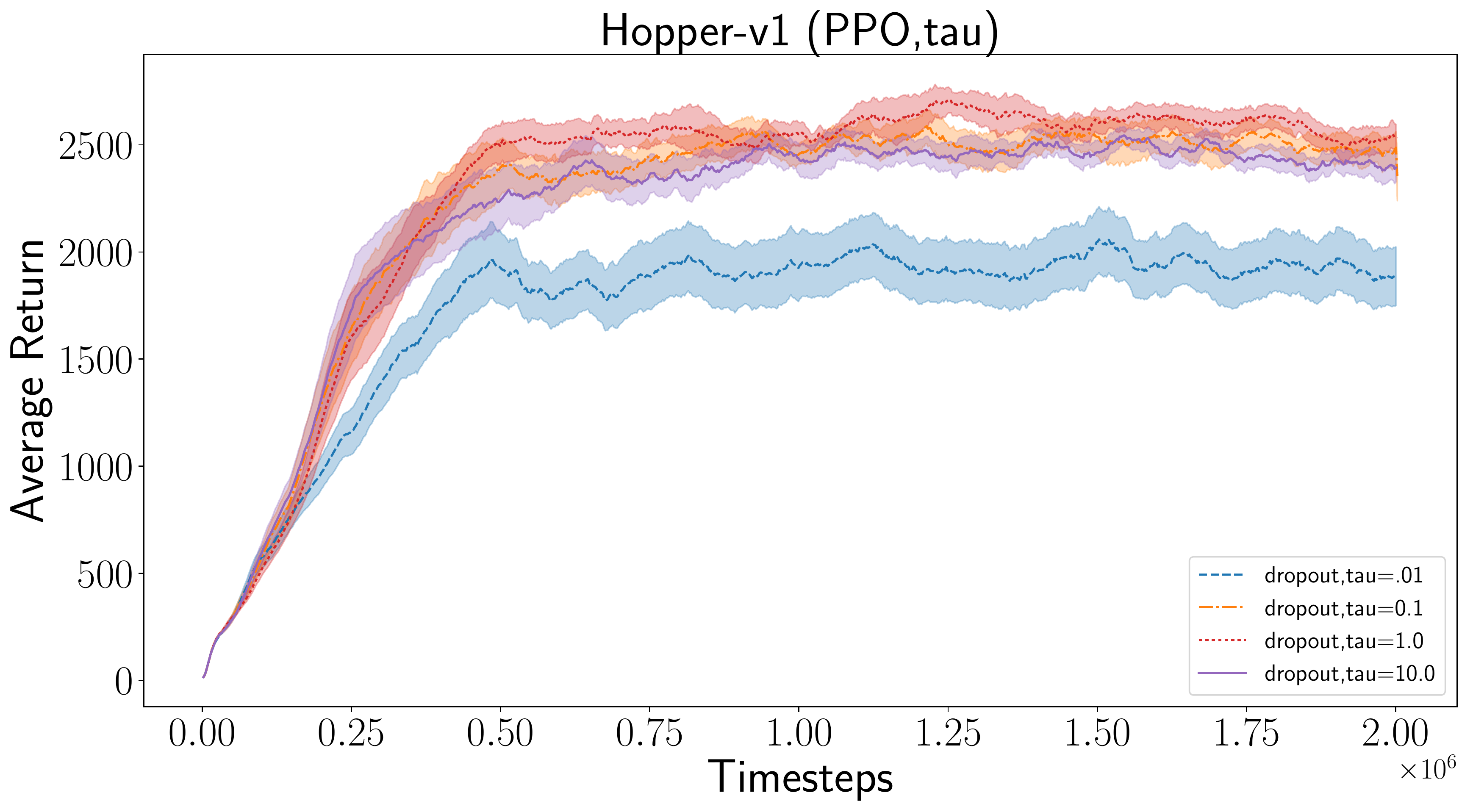}
    \includegraphics[width=.49\textwidth]{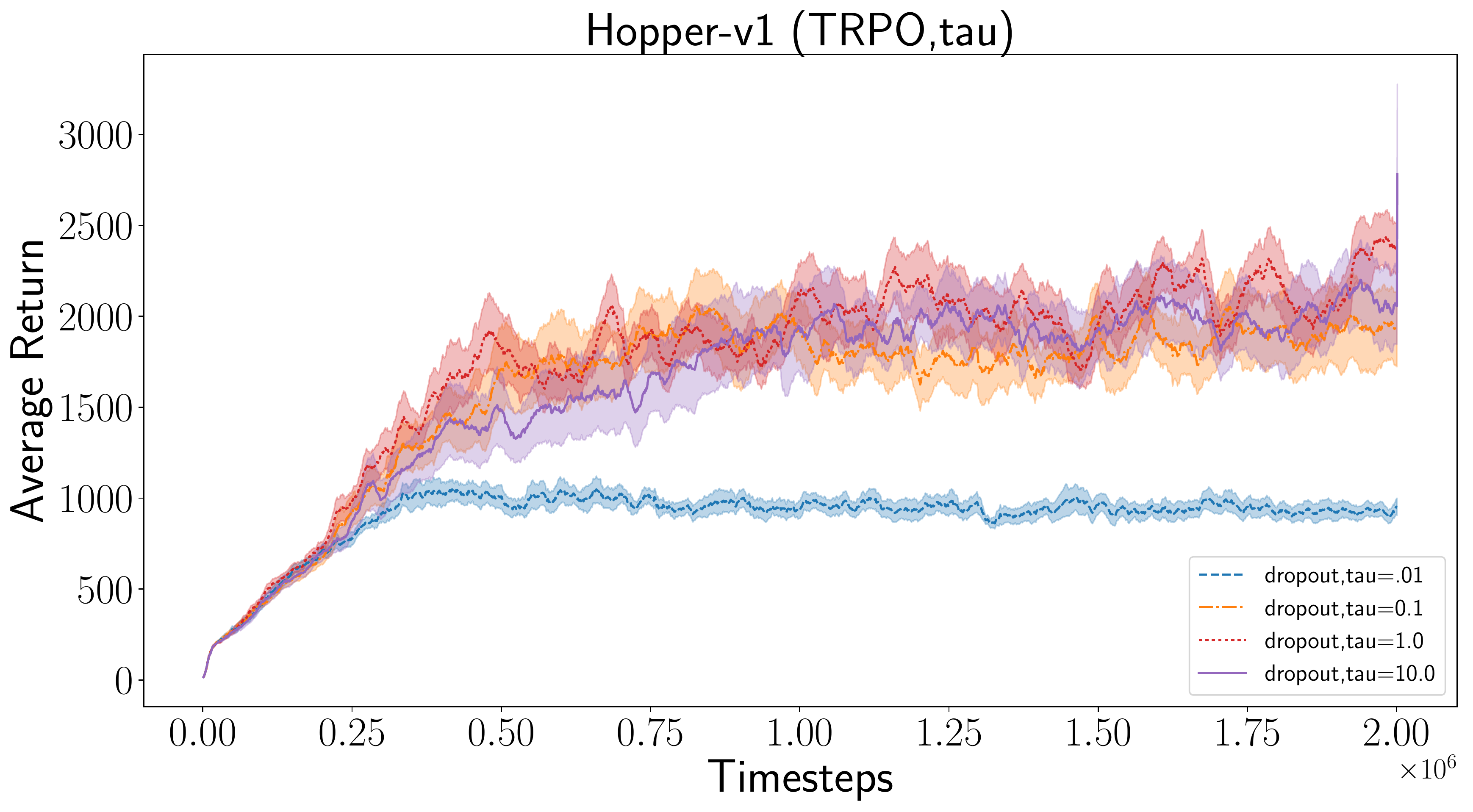}
    \caption{Ablation investigation into the $\tau$ parameter.}
    \label{fig:miniabl}
\end{figure}

We also find that in many cases, the $\alpha$-BNN version held more consistent results across random seeds (lower standard error across trials in Table~\ref{tab:ppo}). While the baseline results see much larger deviations across random seeds, using the $\alpha$-BNN yields fairly consistent results with more stable learning curves (see Appendix for all learning curves).

Finally, we found that the $Q$-value estimates in DDPG were much lower than in the baseline version (see Appendix for more details). We believe this is due to a variance reduction property as in Double-DQN~\cite{DDQN}. While in Double-DQN, two function approximators are used for the $Q$-value function, using dropout effectively creates a parallel to Double-DQN where many different approximators are used from the same set of network weights. This also aligns with~\cite{BootstrappedDQN,bellemare2017distributional}. Thus, as expected, we see a reduction in $Q$-value estimates as in Double-DQN.

\section{Conclusion}

We build off of work in Bayesian Neural Networks~\cite{li2017dropout} and distributional perspectives in uncertain value functions~\cite{bellemare2017distributional} to demonstrate an extension which can successfully be leveraged in continuous control domains with dropout uncertainty estimates. Overall, by providing a simple replacement for value functions, we demonstrate an increase in performance across policy gradient algorithms in continuous control tasks. The significant performance increases in PPO further suggest the importance of a stable baseline which the posterior mean provides.

Our work demonstrates the potential of using Bayesian approximation methods in policy gradient algorithms. This work provides a method which can be used to not only improve performance of existing algorithms by simple replacement of the value function, but be leveraged for more complex uses of value function uncertainty estimates in continuous control. By modelling the uncertainty over value functions in continuous control domains, our work opens up possibilities to use this uncertainty information for other applications such as in safe reinforcement learning.


\section{Acknowledgments}
We would like to thank the AWS Cloud Credits for Research program for compute resources. We'd also like to thank Juan Camilo Gamboa Higuera for helpful discussions.

\bibliographystyle{unsrt}
\bibliography{bibliography}

\newpage
\section{Appendix}
\pdfoutput=1

\subsection{Further Experimental Setup and Results}

\begin{figure}[!htbp]
    \centering
    \includegraphics[width=.49\textwidth]{images/hcppo.pdf}
    \includegraphics[width=.49\textwidth]{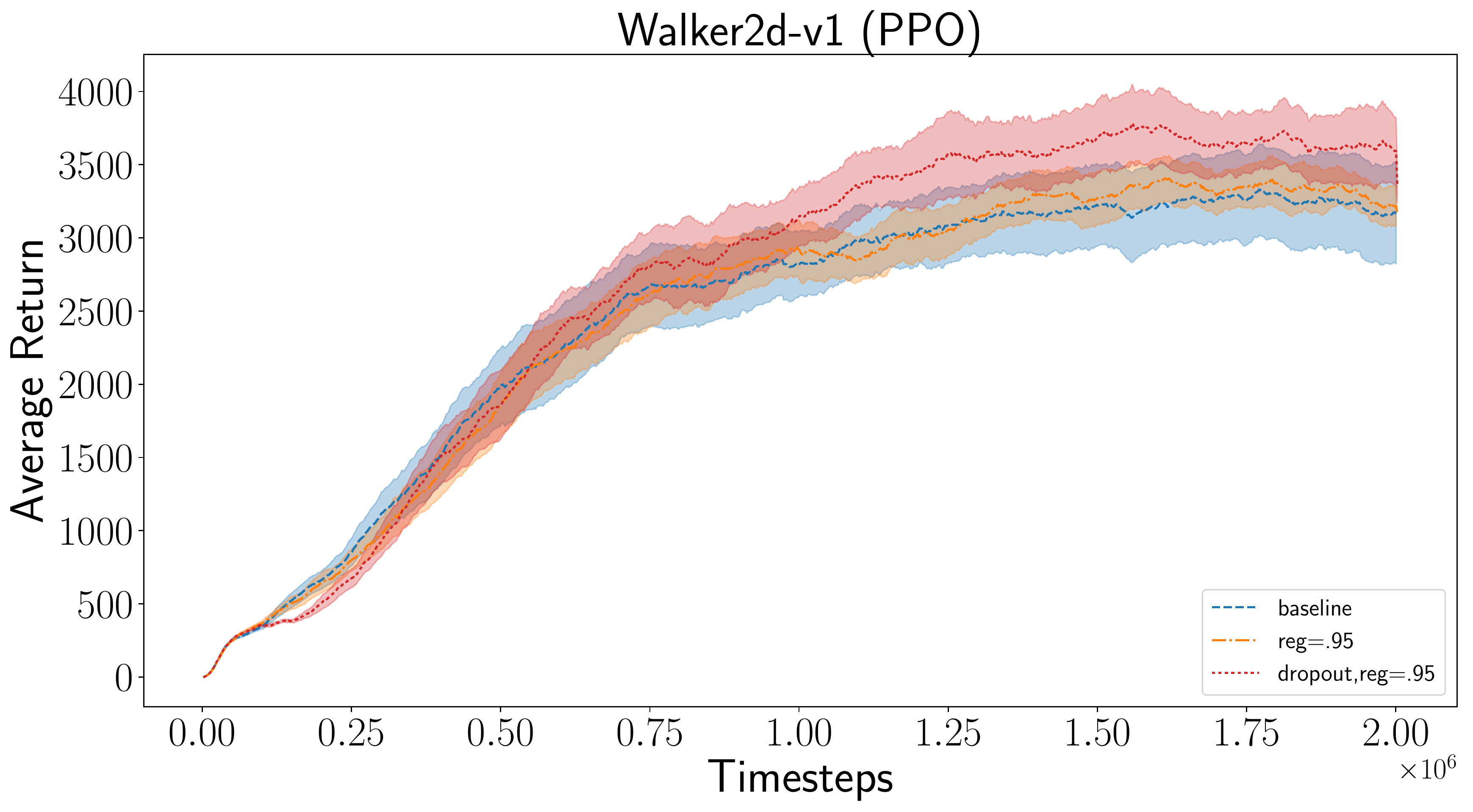}
    \includegraphics[width=.49\textwidth]{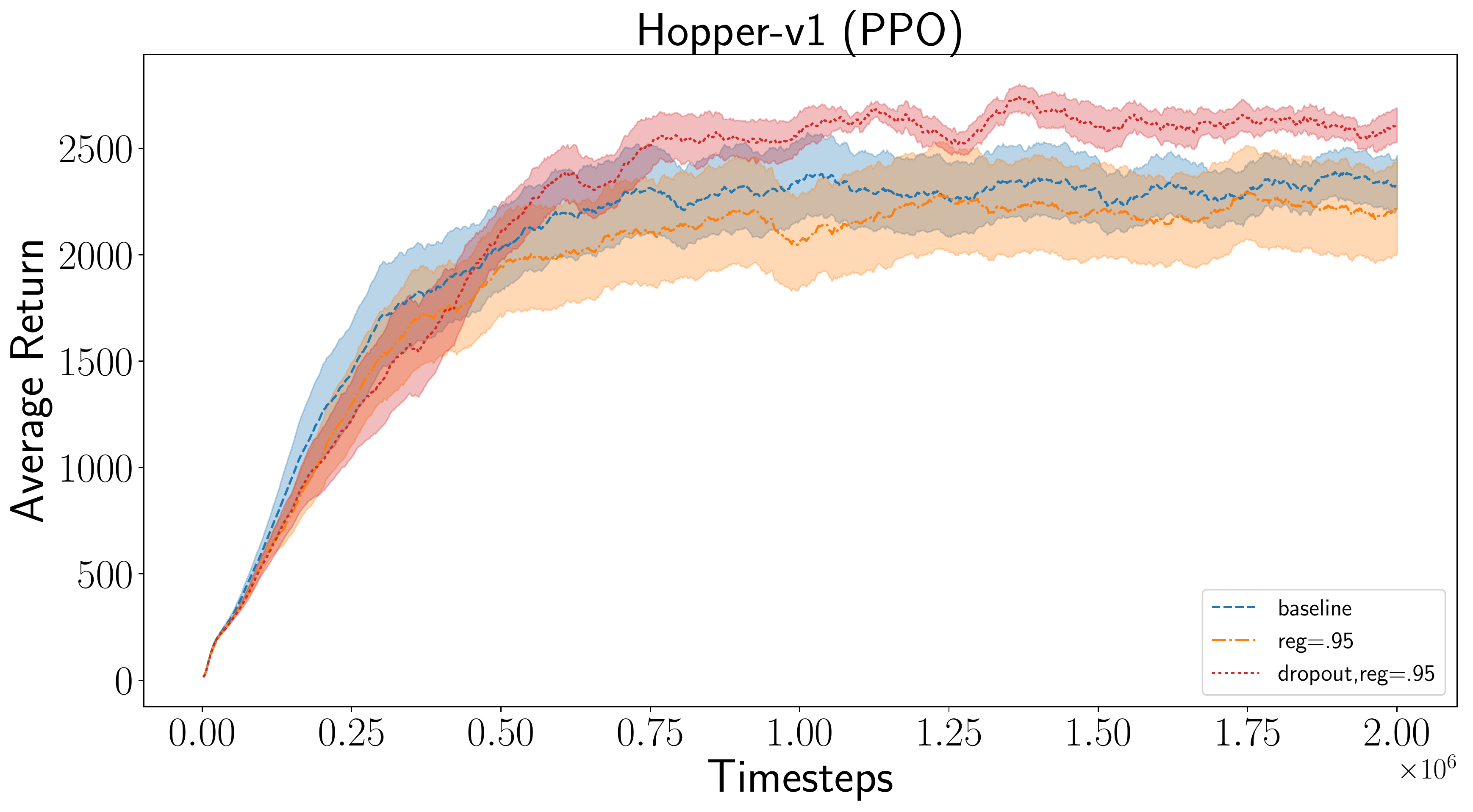}

    \caption{Proximal Policy Optimization with a dropout value function estimator.}
    \label{fig:ppo}
\end{figure}

\begin{figure}[!htbp]
    \centering
    \includegraphics[width=.49\textwidth]{images/hctrpo.pdf}
    \includegraphics[width=.49\textwidth]{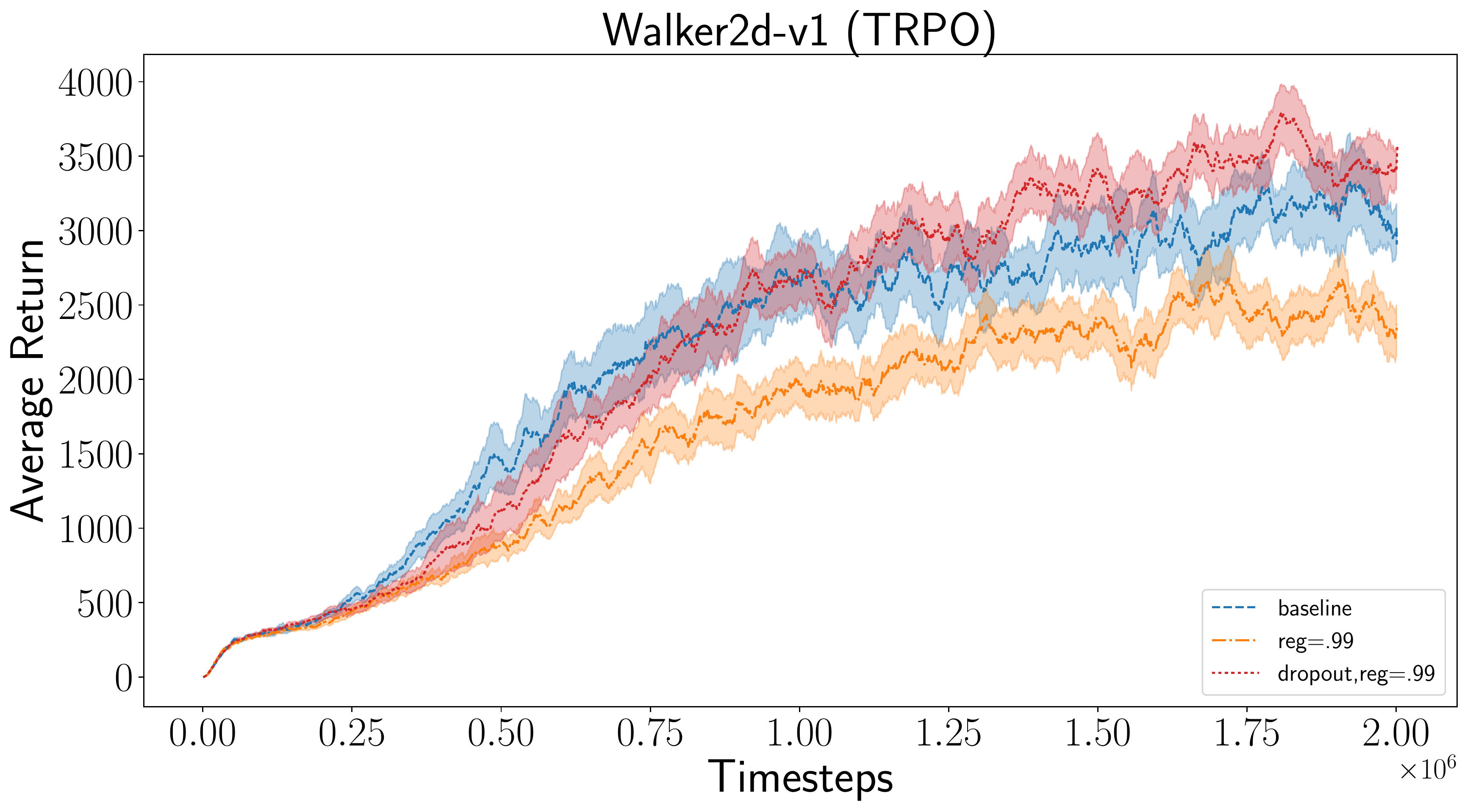}
    \includegraphics[width=.49\textwidth]{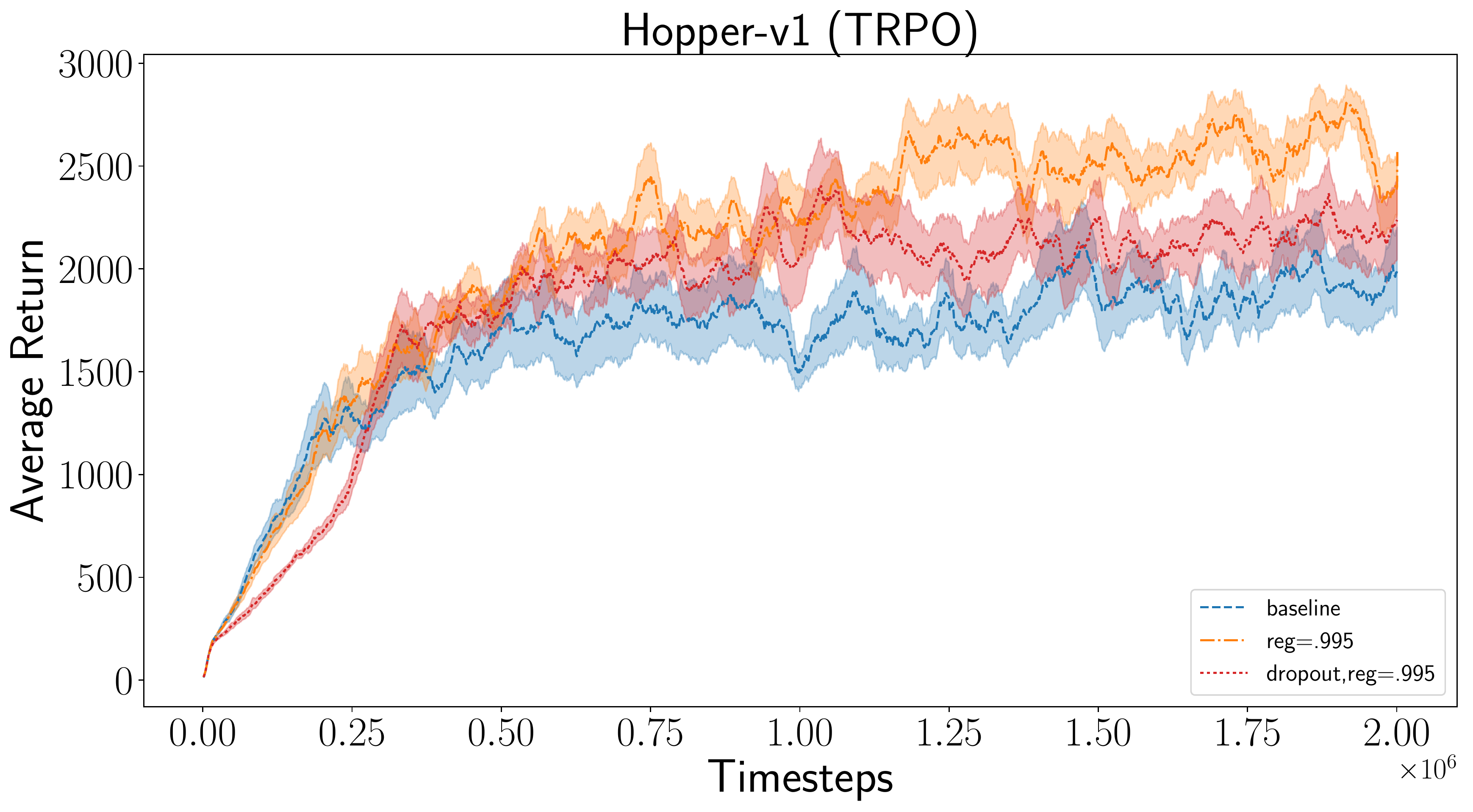}

    \caption{Trust Region Policy Optimization with a dropout value function estimator.}
    \label{fig:trpo}
\end{figure}

For TRPO and PPO experiments we use random seeds $1-10$ inclusive and for DDPG we use seeds $1-5$. We keep all hyperparameters  which are shared between the $\alpha$-BNN version and the baseline version for each algorithm constant. We do not modify hyperparameters from the baselines implementation except that we use \textit{relu} activations instead of \textit{tanh} in the policy and value functions. These can be found in our repository as the default settings in the individual run scripts: \url{https://github.com/Breakend/BayesianPolicyGradients}. For DDPG we use the adaptive exploration strategy from~\cite{plappert2017parameter}. We use 50 Monte Carlo samples in parallel for all experiments. For all optimal experiments we set $\tau=.85$ and as per~\cite{li2017dropout} we set the L2 regularizer to be equivalent to the keep probability on the dropout layers. For PPO we set this keep probability to $.95$ for dropping out network weights and for TRPO and DDPG we set this to $.99$. We use a high dropout keep probability due to the small size of the networks as found via a grid search. See~\cite{gal2016dropout}. We do not use a regularizer (L2) for DDPG as it deteriorated performance. For the $\alpha$-value, we use find that it is only beneficial in TRPO to modify it and set this to $1.0$ in the case of Hopper experiments. All other experiments held $\alpha$ constant at $0.5$. This is according to~\cite{li2017dropout}. This provides a beneficial trade-off between mass covering ($\alpha=1$, KL divergence) and zero forcing ($\alpha = 0$,free energy). Overall, the optimal hyperparameters used were as follows.

For TRPO:

\begin{itemize}
    \item $\alpha=.5$ (Walker,HalfCheetah) $\alpha=1.0$ (Hopper)
    \item $\tau=.85$
    \item $KeepProb=.95$ (HalfCheetah), $KeepProb=.99$ (Walker), $KeepProb=.995$ (Hopper)
    \item $MCSamples=50$
\end{itemize}

For PPO:

\begin{itemize}
    \item $\alpha=.5$
    \item $\tau=.85$
    \item $KeepProb=.75$ (HalfCheetah), $KeepProb=.95$ (Hopper, Walker)
    \item $MCSamples=25$ (Hopper,Walker), $MCSamples=50$ (Hopper,Walker)
\end{itemize}

\subsection{Expanded Analysis}
\pdfoutput=1

Here, we investigate the effects of various parameters and properties of our previously described methods.

\subsubsection{DDPG Q-Value Approximation}
\begin{figure}[!h]
    \centering
    \includegraphics[width=.49\textwidth]{"images/HalfCheetah-v1__DDPG_"}
    \includegraphics[width=.49\textwidth]{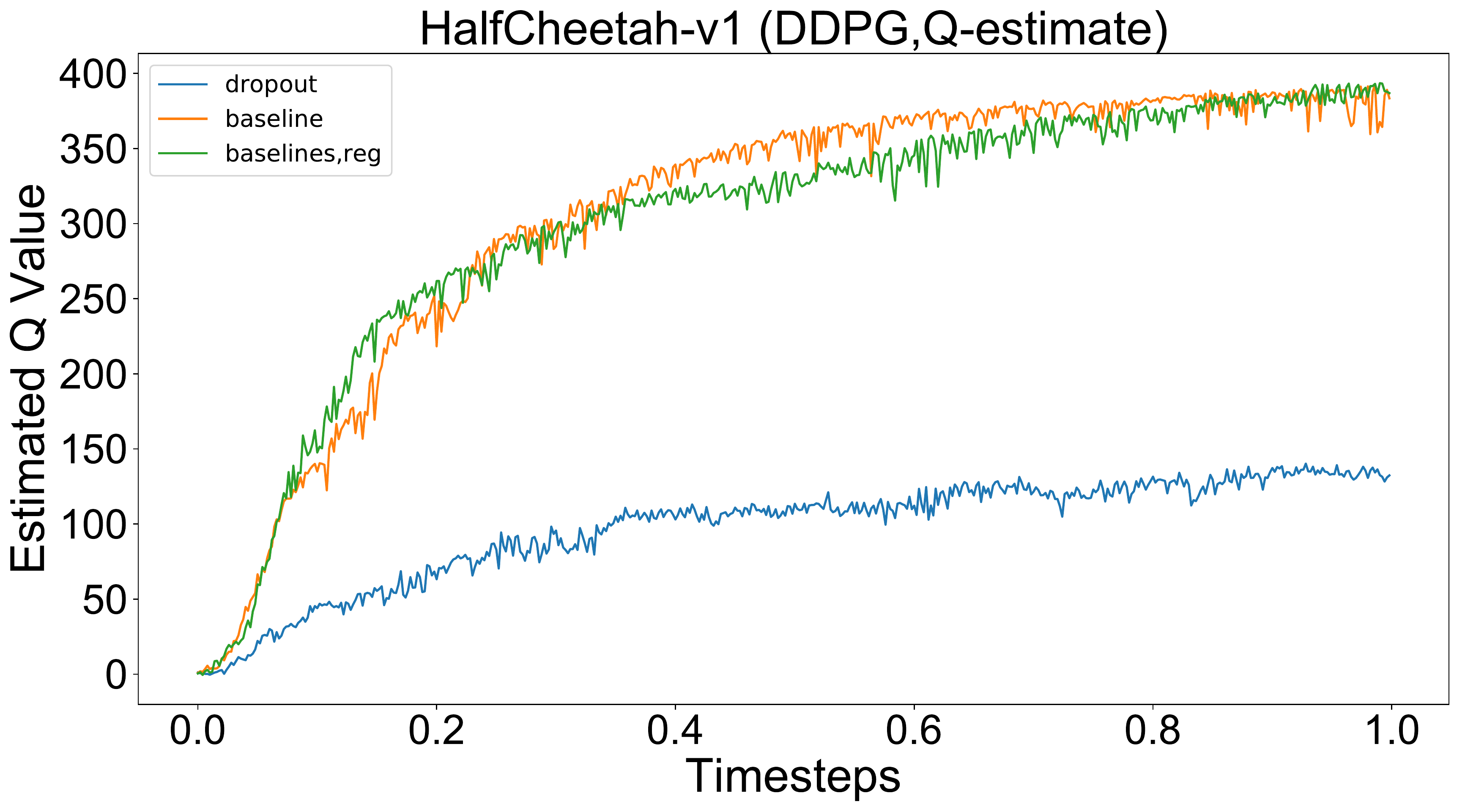}
    \caption{Comparison of Q value estimates during the learning process.}
    \label{fig:qestimate}
\end{figure}
Figure~\ref{fig:qestimate} shows the $Q$-value estimate of the evaluation trials versus the training steps. Similarly to ~\cite{DDQN} (Figure 3 in that work), we notice that the DDPG value estimates are increasing much faster than that of the $\alpha$-BNN value function over time. We suspect that the regularizing property of dropout inference provides a similar variance reduction benefit as in DDQN -- and hence a lower $Q$-estimate overtime -- tackling the overestimation problem in a similar fashion.

\subsubsection{Ablation Analysis}

\begin{figure}[!htbp]
    \centering
    \includegraphics[width=.49\textwidth]{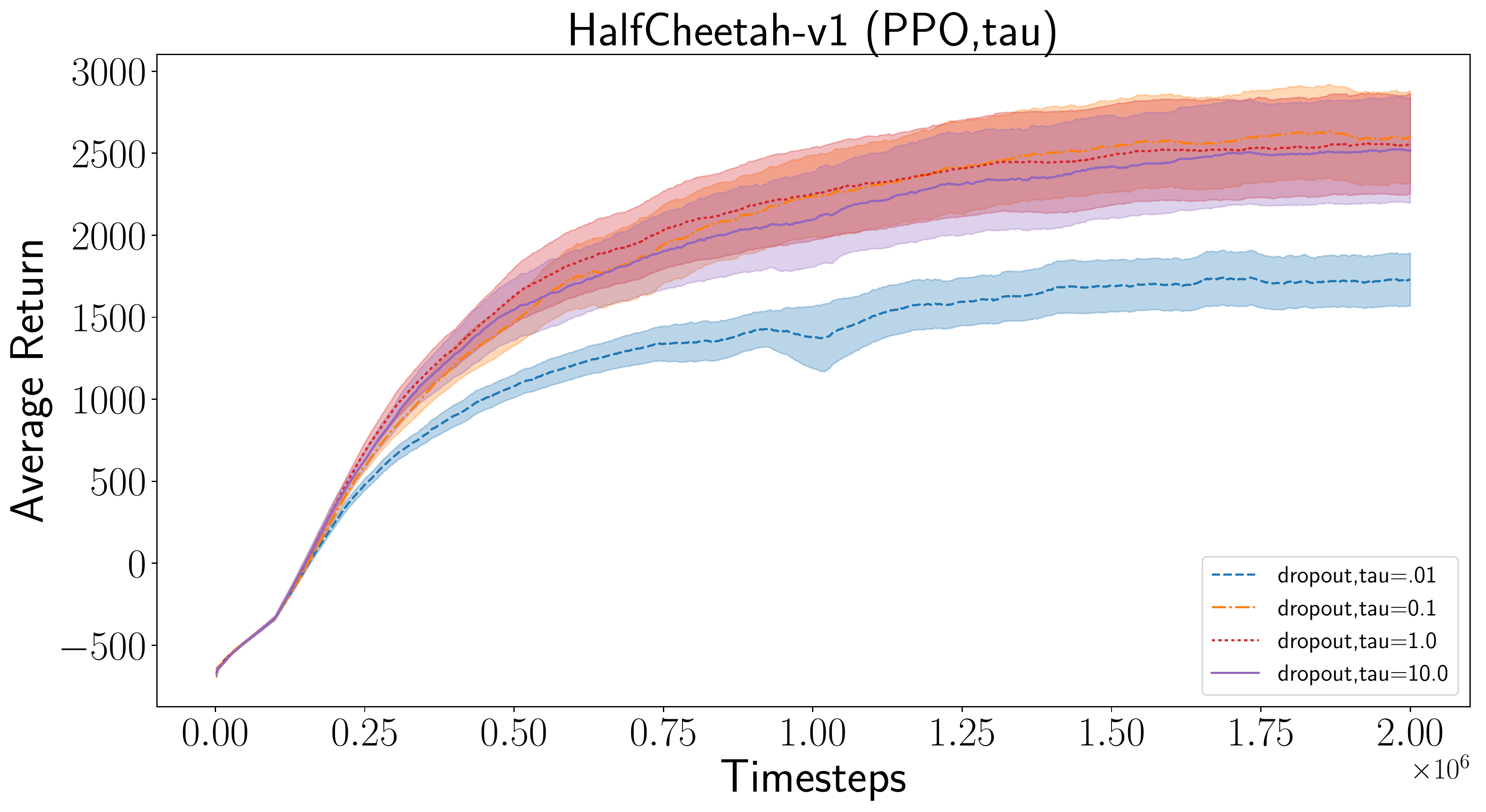}
    \includegraphics[width=.49\textwidth]{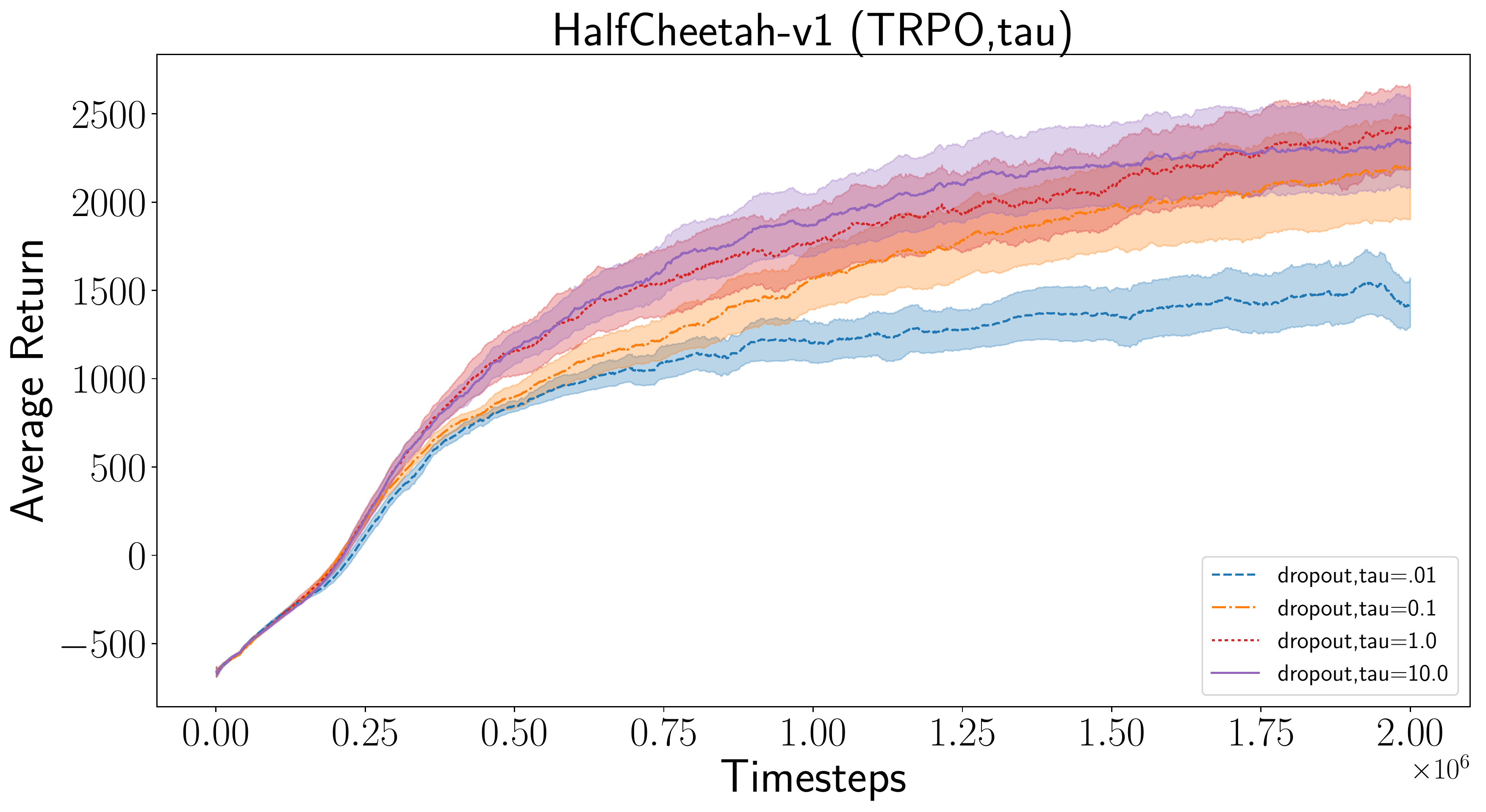}
    \includegraphics[width=.49\textwidth]{"images/Hopper-v1__PPO,tau_"}
    \includegraphics[width=.49\textwidth]{"images/Hopper-v1__TRPO,tau_"}
    \caption{Ablation investigation into the $\tau$ parameter.}
    \label{fig:tau}
\end{figure}

\begin{figure}[!htbp]
    \centering
    \includegraphics[width=.49\textwidth]{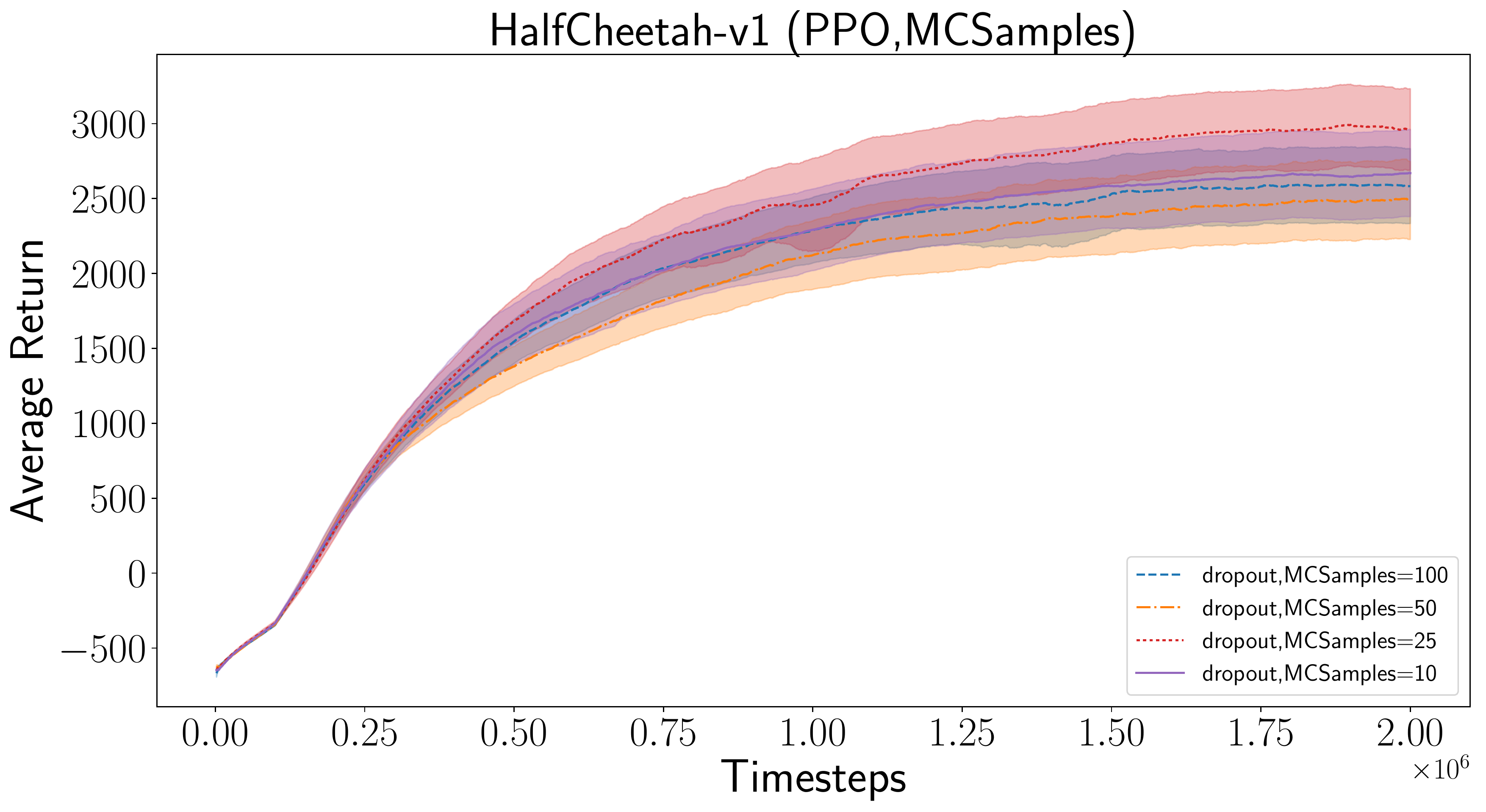}
    \includegraphics[width=.49\textwidth]{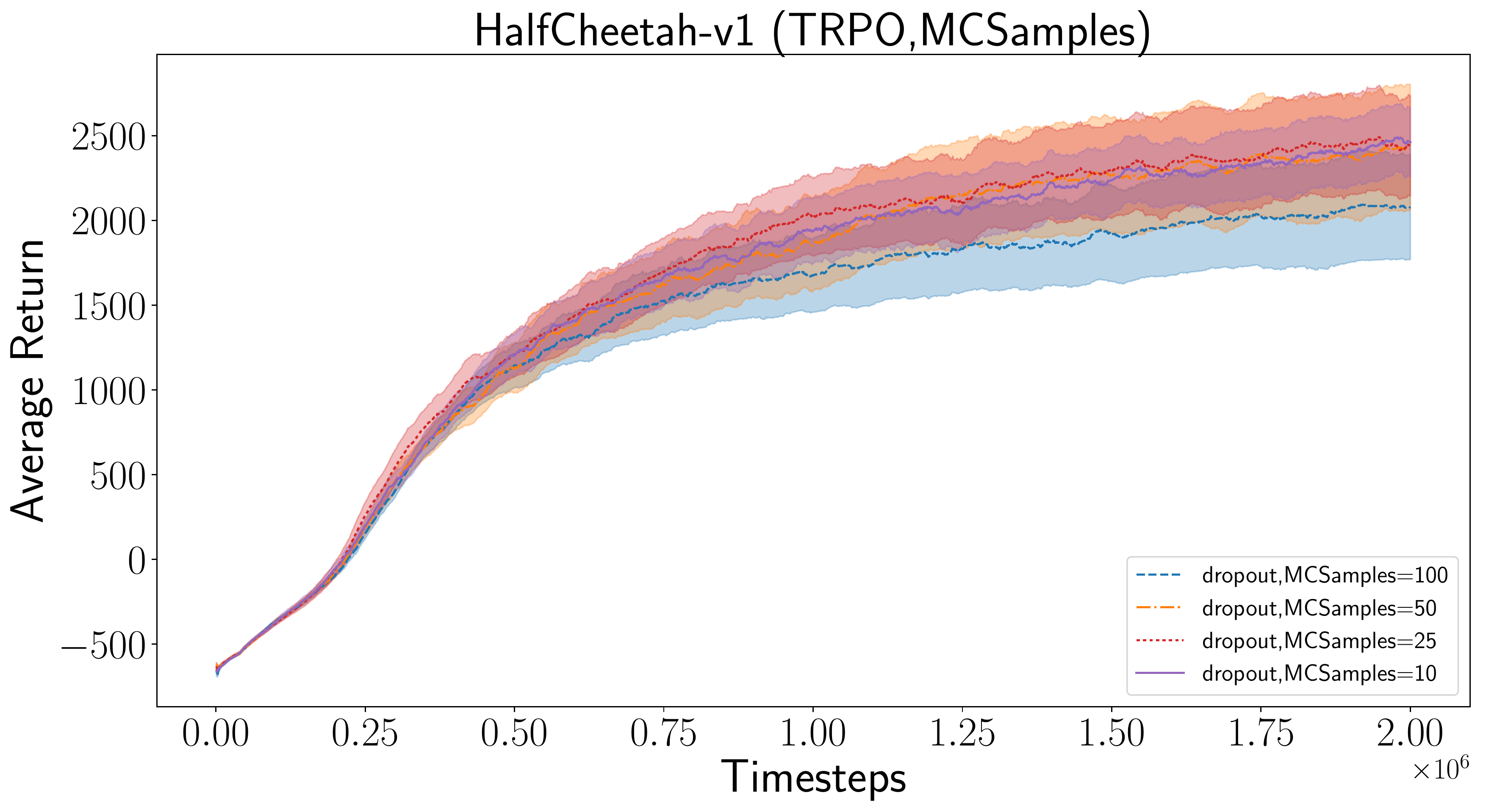}
    \includegraphics[width=.49\textwidth]{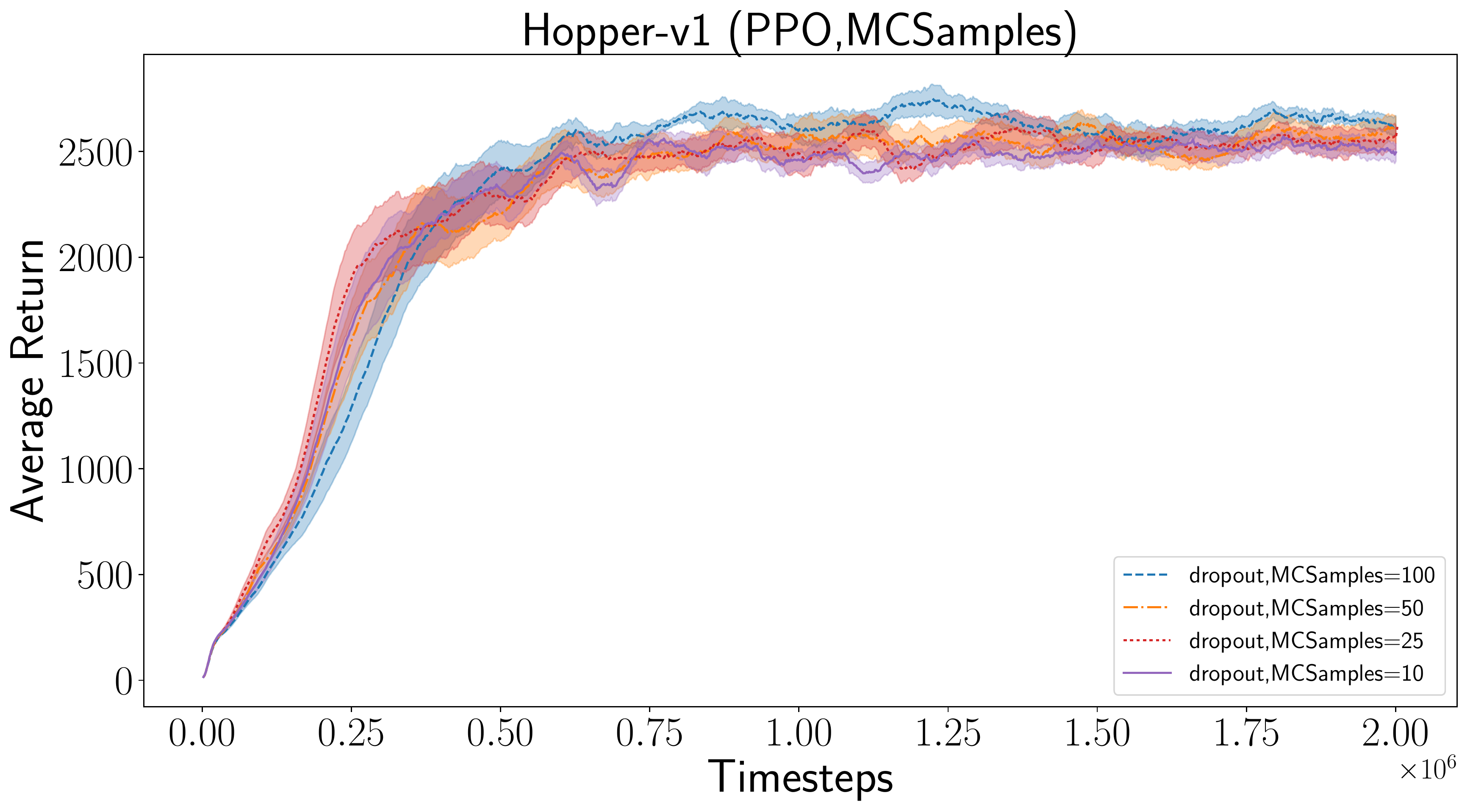}
    \includegraphics[width=.49\textwidth]{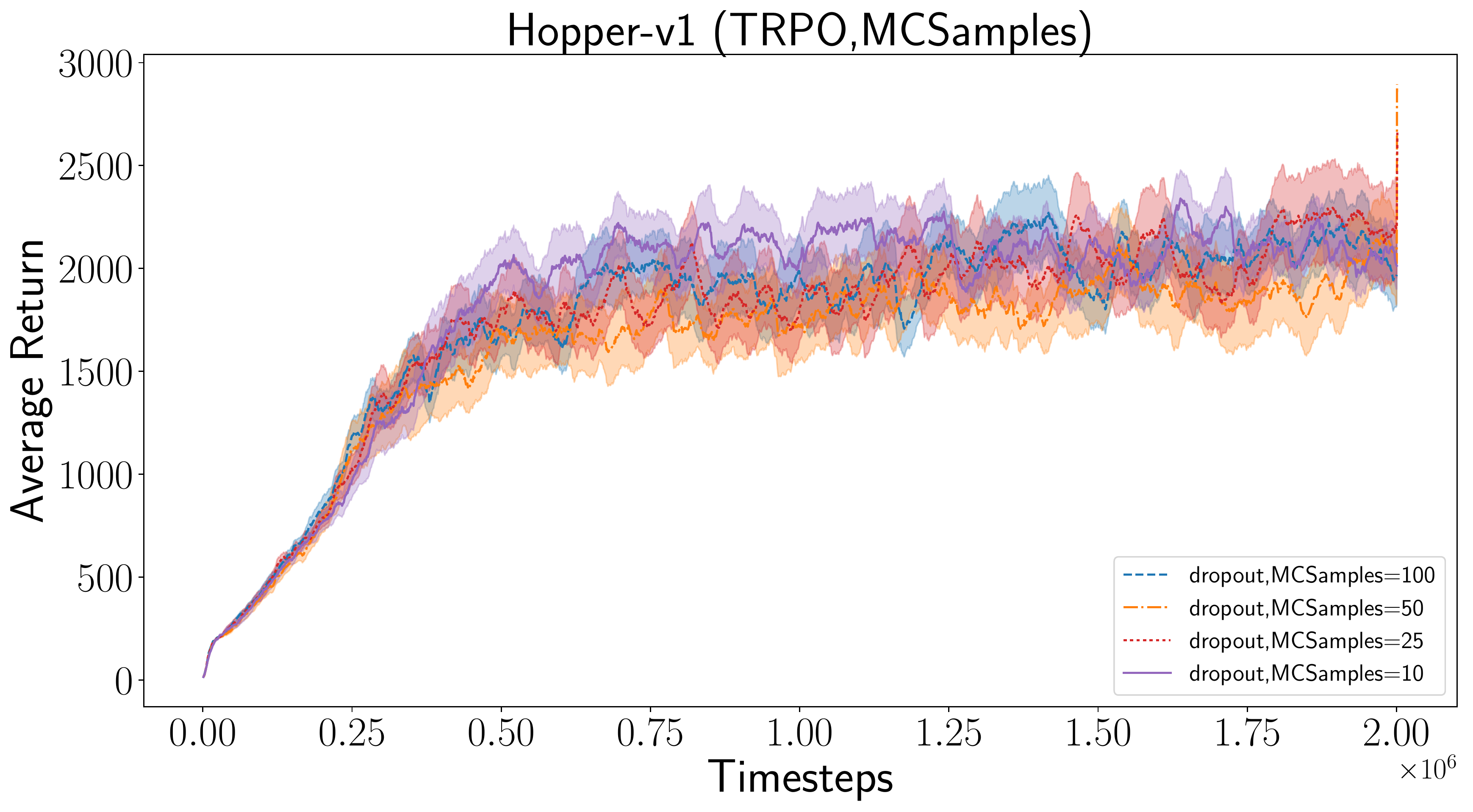}
    \caption{Ablation investigation into the number of MC samples parameter.}
    \label{fig:mc}
\end{figure}

To determine the effect of various hyperparameters, we run ablation analysis on the selection of Hopper and Half-Cheetah environments for TRPO and PPO. We investigate $\tau$, the number of Monte Carlo dropout samples, the $\alpha$ parameter, and the keep probability of the dropout layers. We hold all hyperparameters constant at the default set of $\tau=.85, \alpha=.5, MCSamples=50, KeepProb=.95$. Note that our optimal set of hyperparameters used in the presented main results was found after a subsequent gridsearch across the parameter space.


\begin{figure}[!htbp]
    \centering
    \includegraphics[width=.49\textwidth]{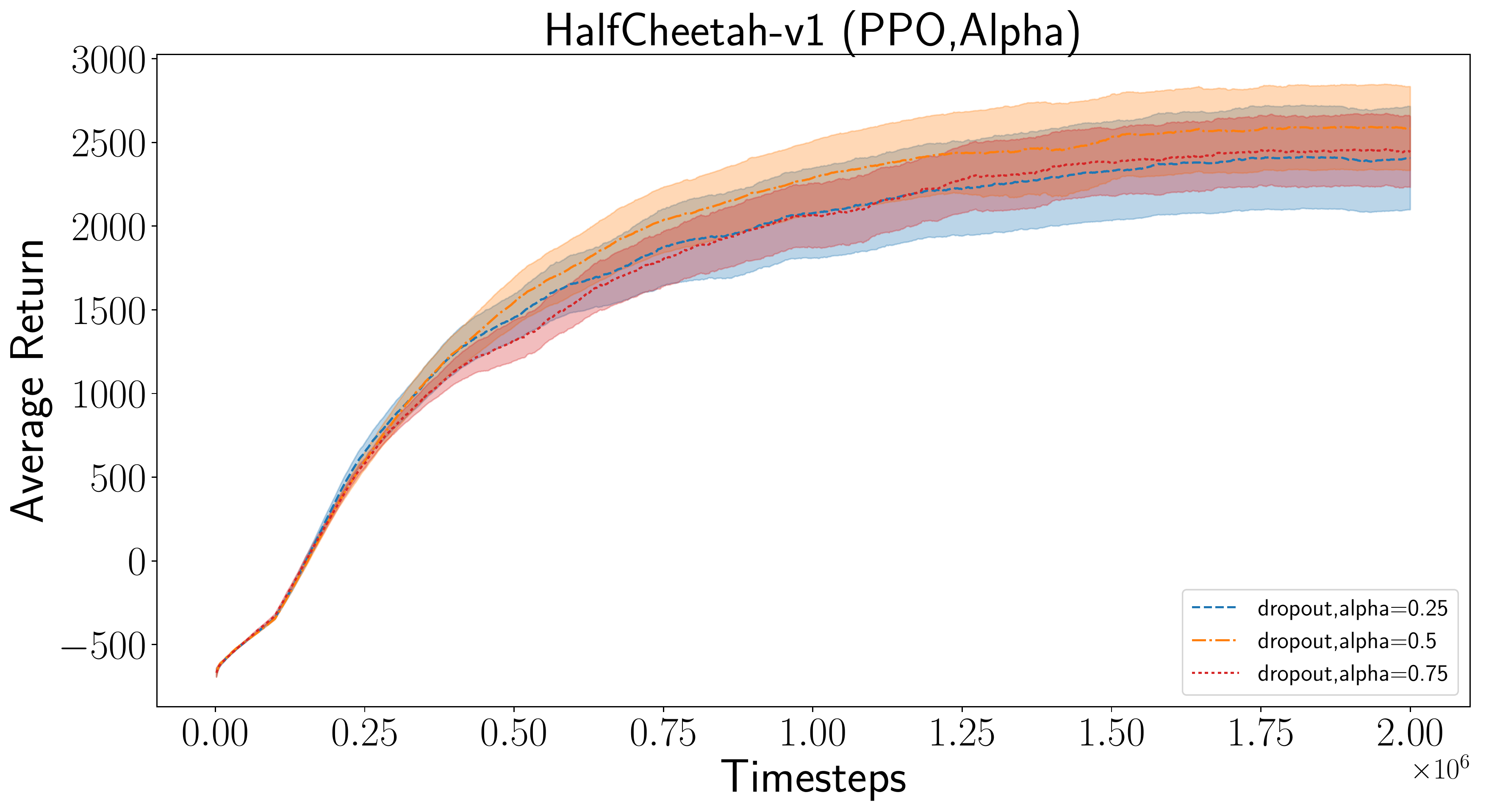}
    \includegraphics[width=.49\textwidth]{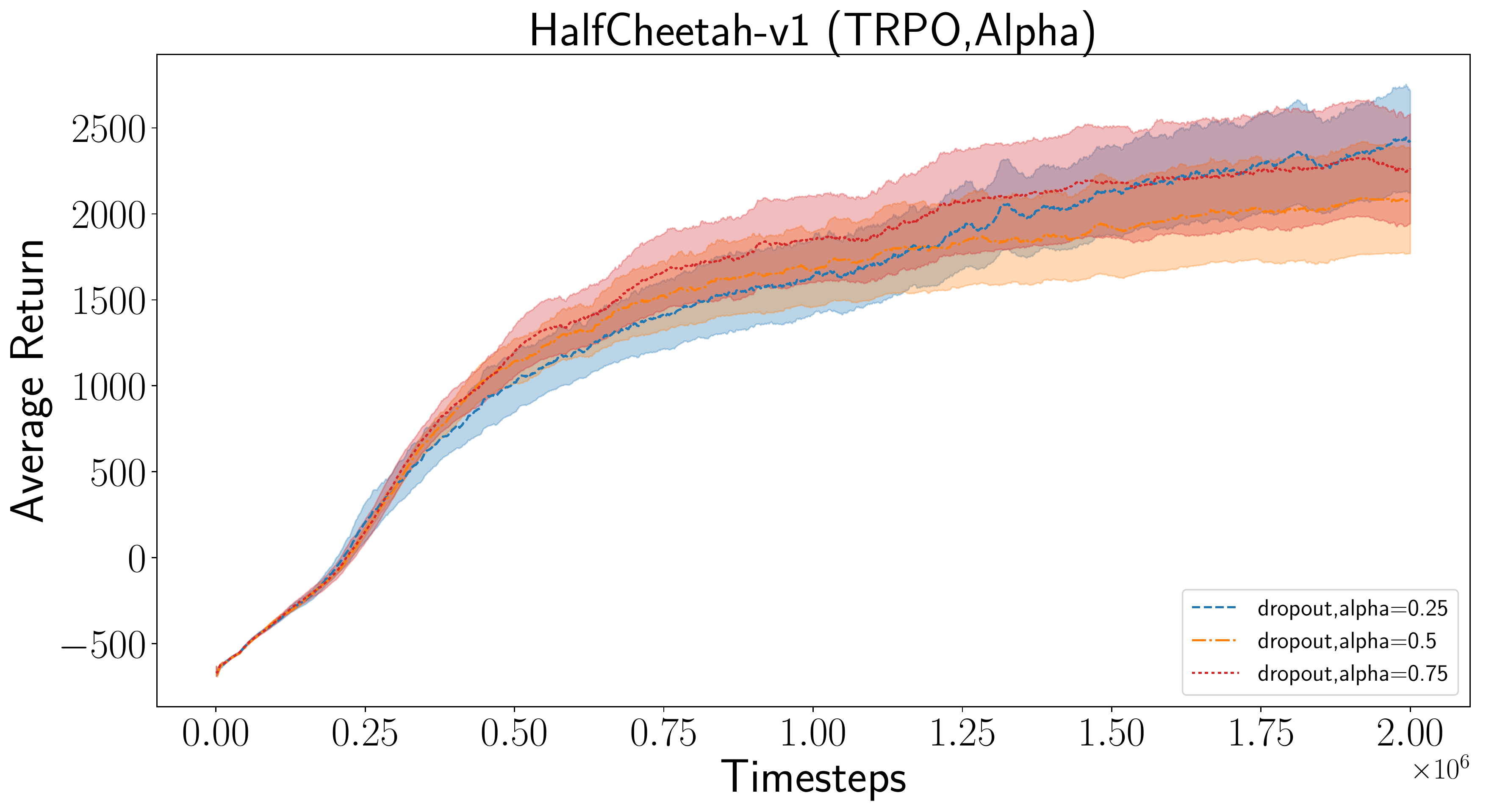}
    \includegraphics[width=.49\textwidth]{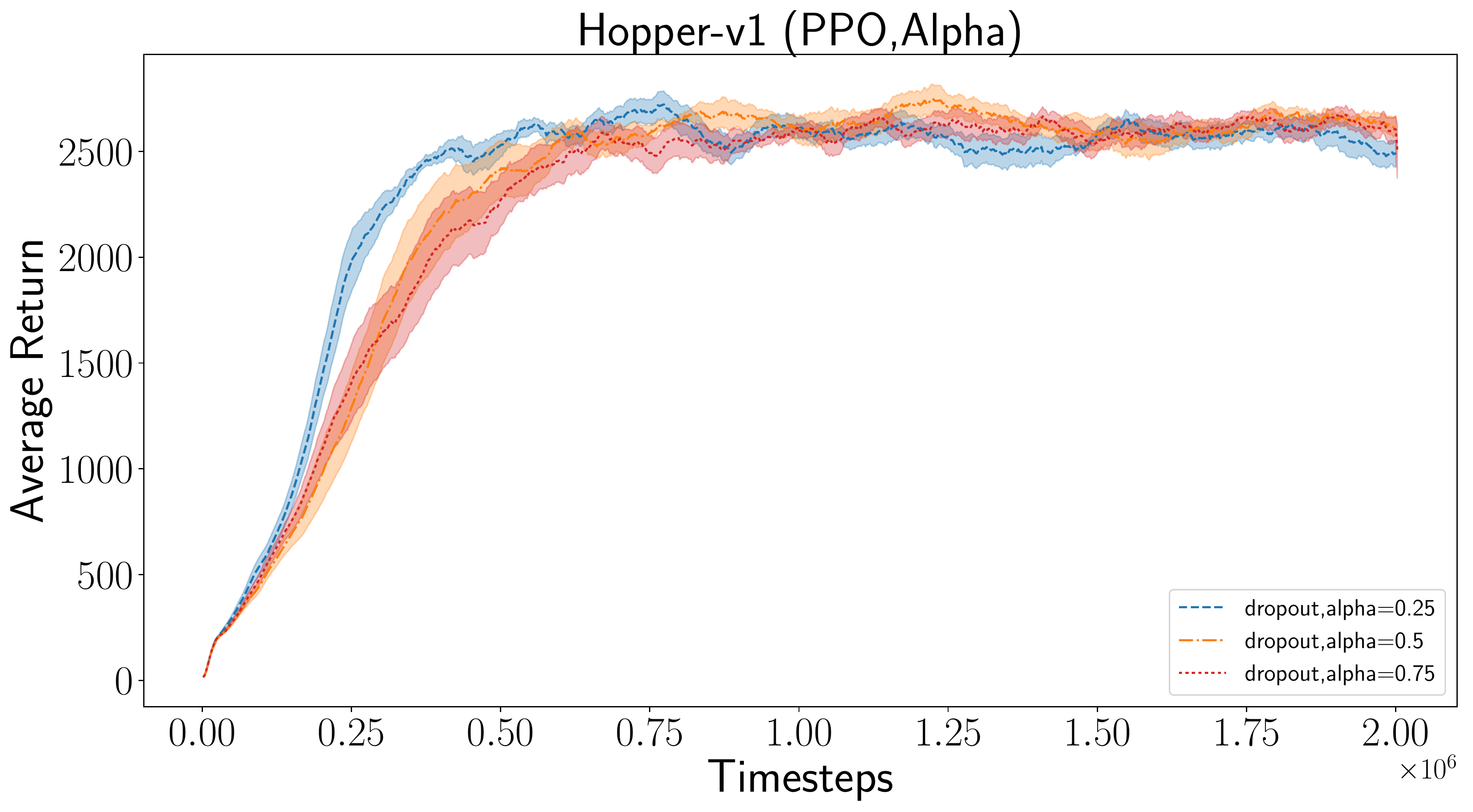}
    \includegraphics[width=.49\textwidth]{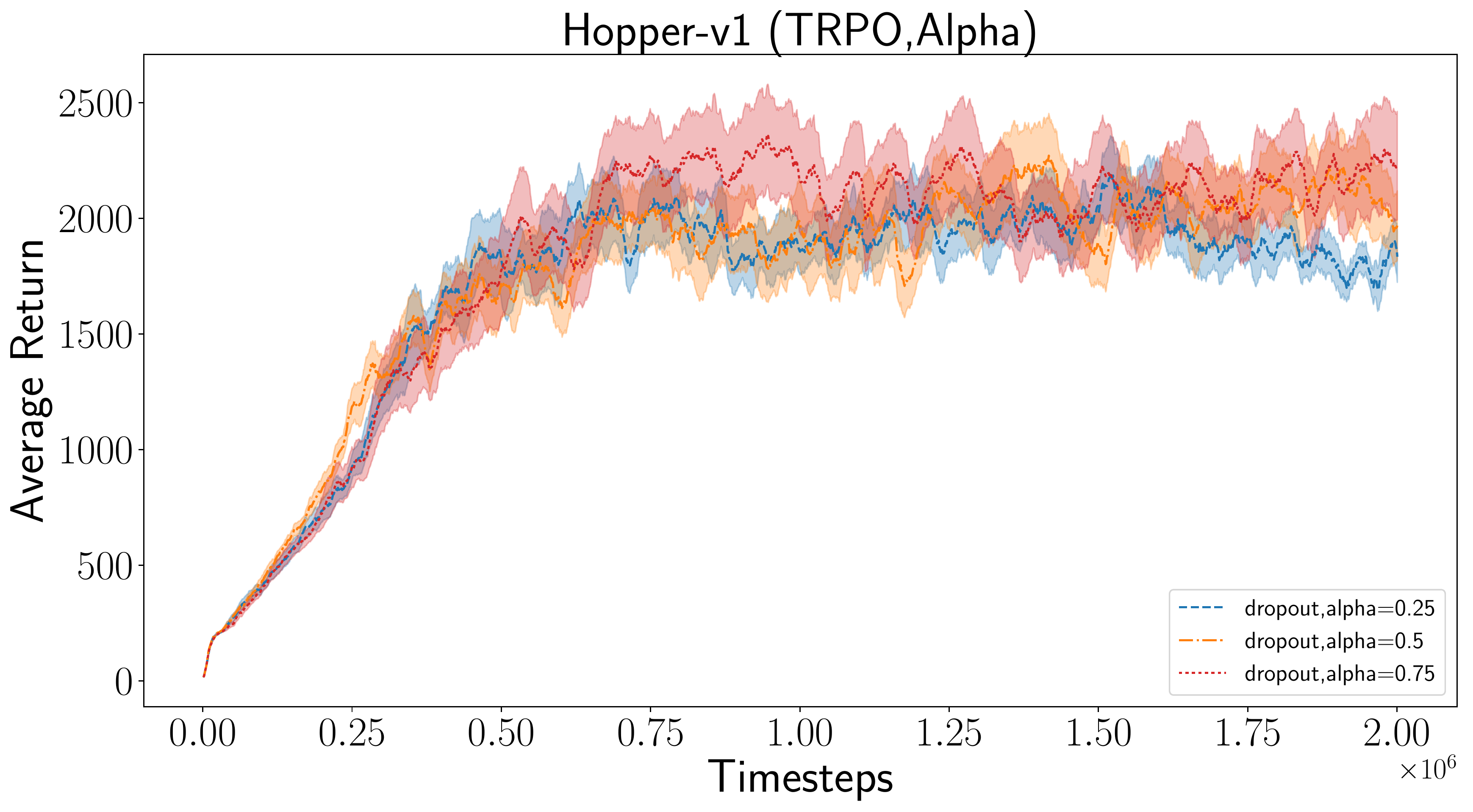}
    \caption{Ablation investigation into the $\alpha$ parameter in the $\alpha$-divergence objective.}
    \label{fig:alpha}
\end{figure}

\begin{figure}[!htbp]
    \centering
    \includegraphics[width=.49\textwidth]{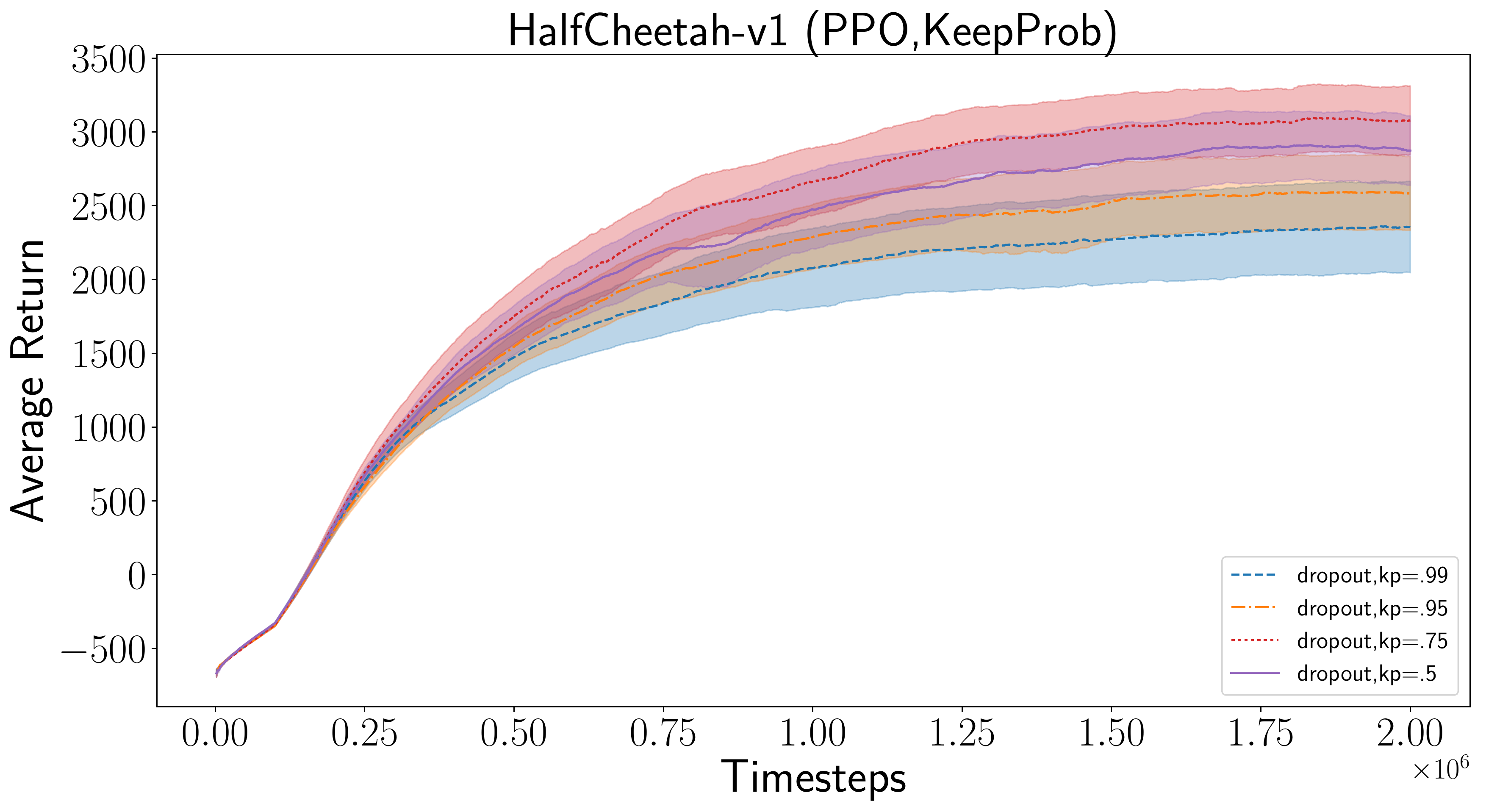}
    \includegraphics[width=.49\textwidth]{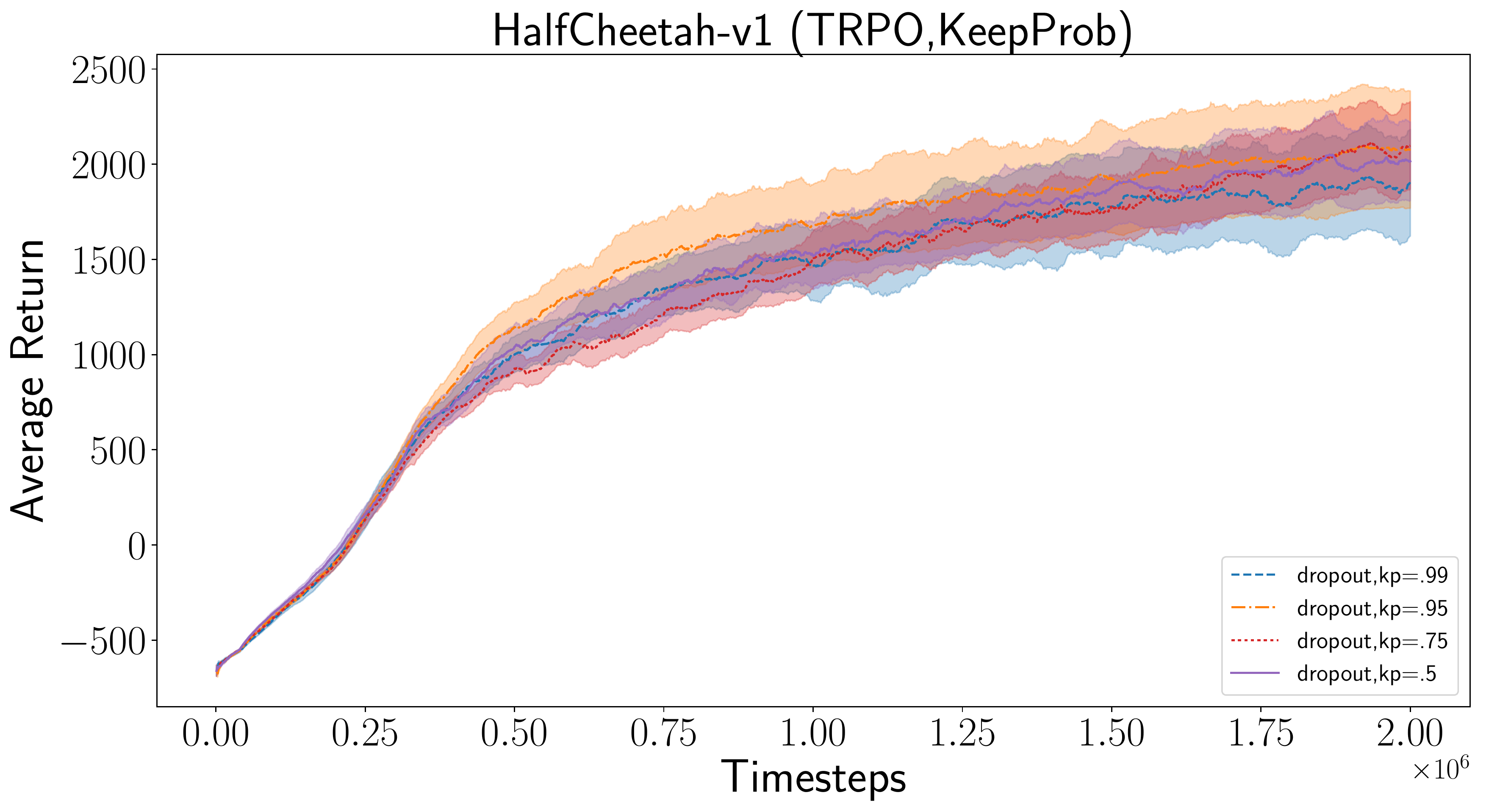}
    \includegraphics[width=.49\textwidth]{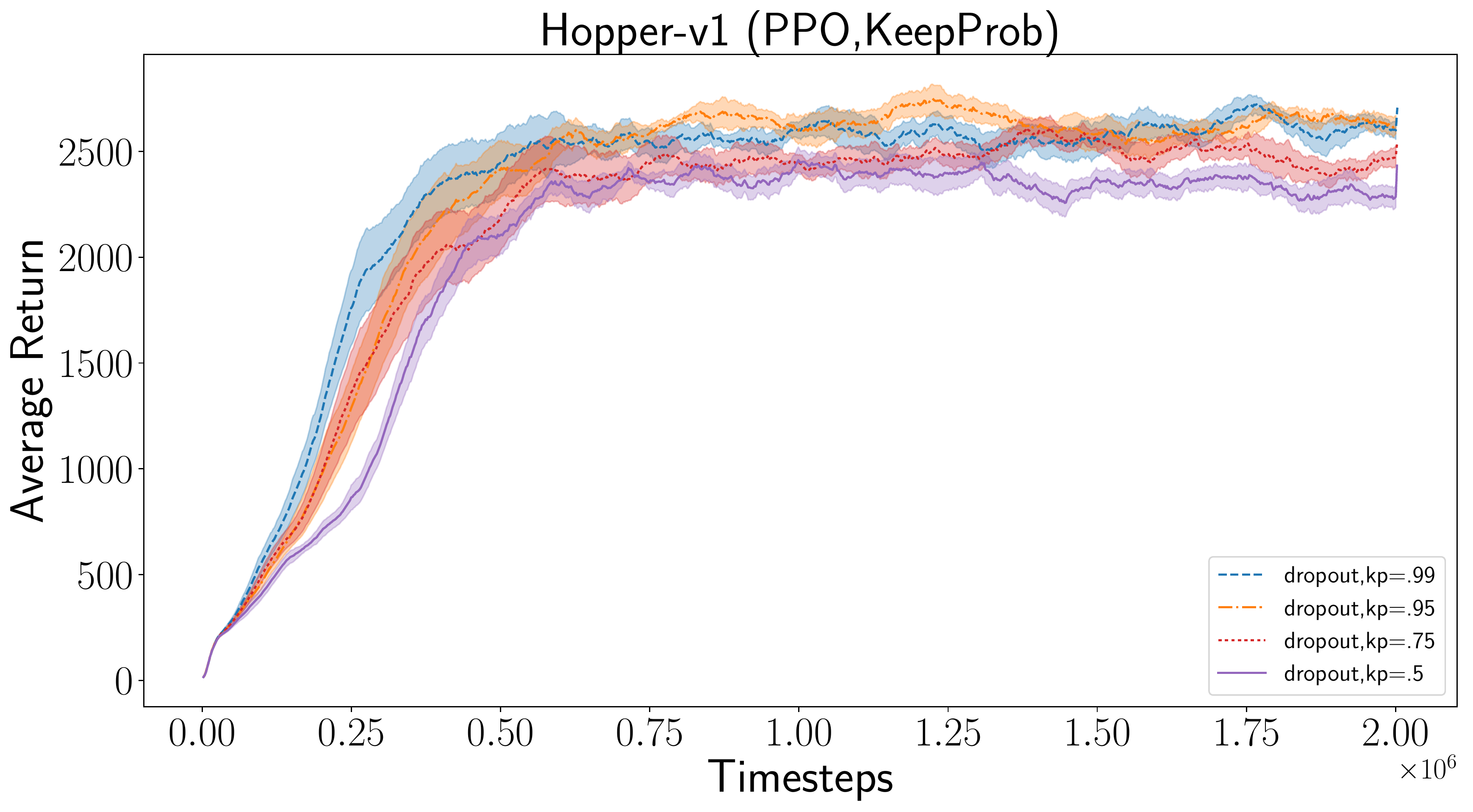}
    \includegraphics[width=.49\textwidth]{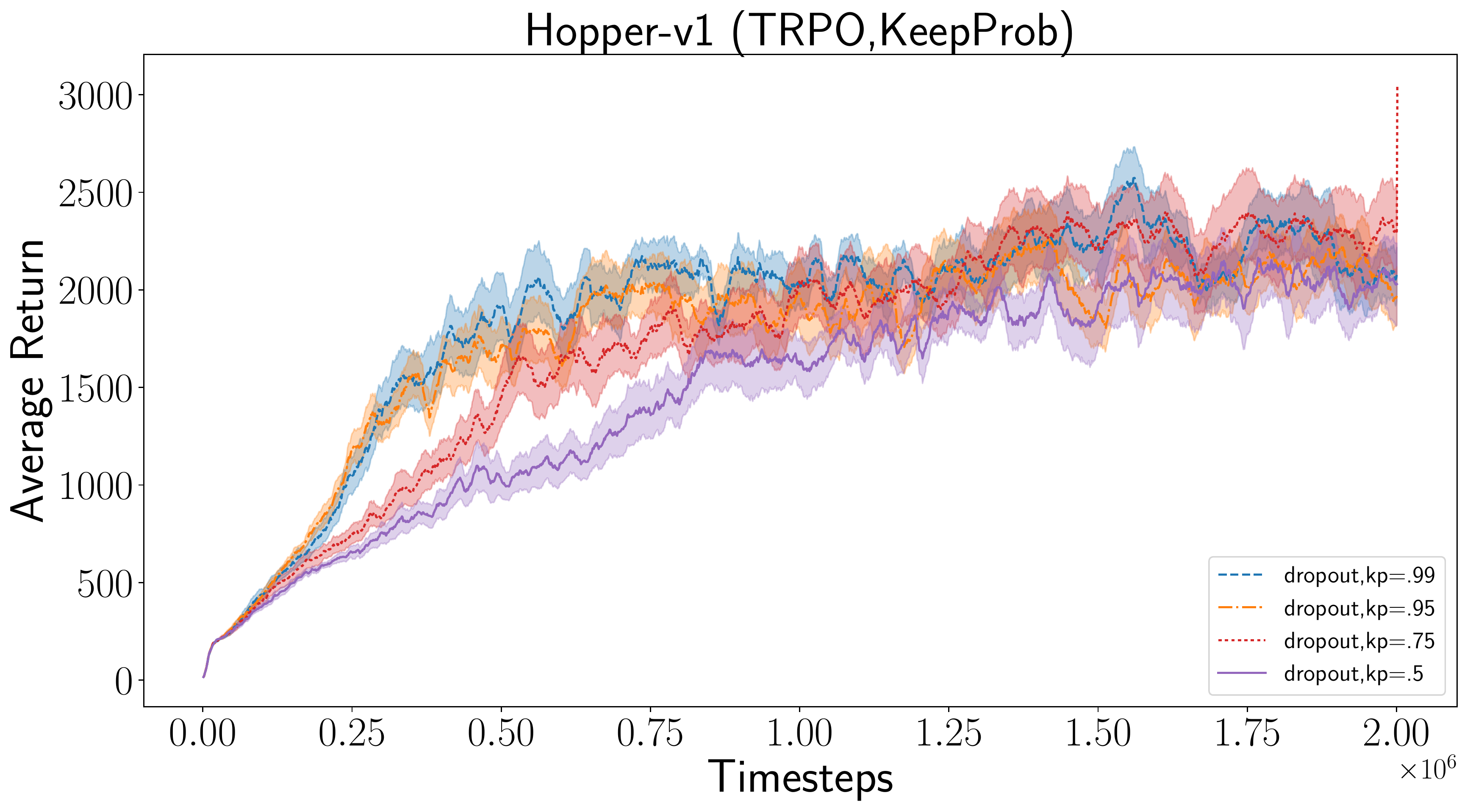}
    \caption{Ablation investigation into the keep probability for the dropout function.}
    \label{fig:keepprob}
\end{figure}

Figures~\ref{fig:tau} shows ablation analysis where we vary the $\tau$ hyperparameter. Overall, we find that HalfCheetah-v1 is more susceptible to variation due to hyperparameter changes. This is likely due to the stable nature of this environment over others like Hopper-v1. This is discussed in~\cite{rlmatters}. The $\tau$ parameter can be thought of as the trade-off between optimization of the objective loss and regularization. As such we can see large variations in performance due to emphasis on the different components. Furthermore, Figures~\ref{fig:keepprob},~\ref{fig:mc},~\ref{fig:alpha} investigate the $KeepProbability$, $MCSamples$, and $\alpha$ paramater of the $\alpha$-divergence objective. We find that in some cases (e.g. as with the $KeepProbability$ in PPO HalfCheetah) effects vary significantly and can cause performance gains, while in others generally do not affect results (e.g. with $\alpha$ in PPO). We find also that effects of hyperparameters vary between PPO and TRPO, showing that despite them both using trust regions, these two algorithms have different properties and dynamics.

\subsection{Extended Proof for Alpha Divergence DDPG Updates }

To demonstrate how the Monte Carlo dropout distributional expectation fits into the DDPG update, we can examine the following proof as per the original DDPG derivation. We want to prove that by maximizing the posterior reward $\mathbb{E}[J(\mu_{\theta})|\mathcal{D}]$,
we end up with the following Bayesian DDPG updates:\\

$$\nabla_{\theta}\mathbb{E}[J(\mu_{\theta})|\mathcal{D}]=\mathbb{E}_{s\sim\rho,\omega\sim q(\omega)}[\nabla_{\theta}\mu_{\theta}(s)\nabla_{a}Q^{\pi}(s,a;\omega)|_{a=\mu_{\theta}(s)}]$$

where\\

$$\mathbb{E}[J(\mu_{\theta})|\mathcal{D}]=\int_{S}p_{1}(s)\int_{\omega}q(\omega)V^{\mu_{\theta}}(s)d\omega ds=\int_{S}p_{1}(s)\int_{\omega}q(\omega)Q^{\mu_{\theta}}(s,\mu_{\theta}(s))d\omega ds$$

Proof, notations and framework are largely adapted from~\cite{DDPG},
please refer to it for more details.\\
$Q^{\mu_{\theta}}(s,\mu_{\theta}(s))=Q^{\mu_{\theta}}(s,\mu_{\theta}(s);\omega)$ where $\omega$ are the weight of $Q(s,a)$\\
\begin{align*}
\mathbb{E}_{\omega\sim q(\omega)}[\nabla_{\theta}V^{\mu_{\theta}}(s)] & =\nabla_{\theta}\mathbb{E}_{\omega\sim q(\omega)}[Q^{\mu_{\theta}}(s,\mu_{\theta}(s))]\\
 & =\nabla_{\theta}(r(s,\mu_{\theta}(s)+\int_{S}\gamma p(s'|s,\mu_{\theta}(s))\int_{\omega}q(\omega)Q^{\mu_{\theta}}(s,\mu_{\theta}(s'))d\omega ds')\\
 & =\nabla_{\theta}\mu_{\theta}(s)\nabla_{a}r(s,a)|_{a=\mu_{\theta}(s)}+\int_{S}\gamma\nabla_{\theta}\mu_{\theta}(s)\nabla_{a}p(s'|s,a)|_{a=\mu_{\theta}(s)}\int_{\omega}q(\omega)Q^{\mu_{\theta}}(s',\mu_{\theta}(s'))d\omega ds'\\
 & +\int_{S}\gamma p(s'|s,\mu_{\theta}(s))\int_{\omega}q(\omega)\nabla_{\theta}V^{\mu_{\theta}}(s')d\omega ds'\\
 & =\nabla_{\theta}\mu_{\theta}(s)\nabla_{a}(r(s,a)+\int_{S}\gamma p(s'|s,a)\int_{\omega}q(\omega)Q^{\mu_{\theta}}(s,\mu_{\theta}(s'))d\omega ds')|_{a=\mu_{\theta}(s)}\\
 & +\int_{S}\gamma p(s'|s,\mu_{\theta}(s))\nabla_{\theta}\int_{\omega}q(\omega)Q^{\mu_{\theta}}(s,\mu_{\theta}(s'))d\omega ds'\\
 & =\underbrace{\nabla_{\theta}\mu_{\theta}(s)\nabla_{a}\mathbb{E}_{\omega\sim q(\omega)}[Q^{\mu_{\theta}}(s,\mu_{\theta}(s))]|_{a=\mu_{\theta}(s)}}_{1}\\
 & +\underbrace{\int_{S}\gamma p(s\rightarrow s',1,\mu_{\theta})\nabla_{\theta}\mathbb{E}_{\omega\sim q(\omega)}[Q^{\mu_{\theta}}(s',\mu_{\theta}(s'))]ds}_{2}
\end{align*}
Expanding $Q^{\mu_{\theta}}(s,\mu_{\theta}(s'))$ in the 2nd term and iterating gives:\\
\begin{align*}
\mathbb{E}_{\omega\sim q(\omega)}[\nabla_{\theta}V^{\mu_{\theta}}(s)] & =1+\int_{S}\gamma p(s\rightarrow s',1,\mu_{\theta})\nabla_{\theta}\mu_{\theta}(s)\nabla_{a}\mathbb{E}_{\omega\sim q(\omega)}[Q^{\mu_{\theta}}(s',\mu_{\theta}(s'))]|_{a=\mu_{\theta}(s')}ds'\\
 & +\int_{S}\gamma^{2}p(s\rightarrow s',1,\mu_{\theta})\int_{S}p(s'\rightarrow s'',1,\mu_{\theta})\nabla_{\theta}\mathbb{E}_{\omega\sim q(\omega)}[Q^{\mu_{\theta}}(s,\mu_{\theta}(s''))]ds''ds'\\
 & ...\\
= & \int_{S}{\displaystyle {\displaystyle \sum_{t=0}^{\infty}\gamma^{t}p(s\rightarrow s',t,\mu_{\theta})\nabla_{\theta}\mu_{\theta}(s')\nabla_{a}\mathbb{E}_{\omega\sim q(\omega)}[Q^{\mu_{\theta}}(s,\mu_{\theta}(s'))]|_{a=\mu_{\theta}(s')}ds'}}
\end{align*}
\\
Finally, we get:\\
\begin{align*}
\nabla_{\theta}\mathbb{E}_{s\sim\rho,a\sim\mu_{\theta},\omega\sim q(\omega)}[J(\mu_{\theta})]= & \nabla_{\theta}\int_{S}p_{1}(s)\int_{\omega}q(\omega)Q^{\mu_{\theta}}(s,\mu_{\theta}(s))d\omega ds\\
= & \mathbb{E}_{s\sim\rho,\omega\sim q(\omega)}[\nabla_{\theta}\mu_{\theta}(s)\nabla_{a}Q^{\pi}(s,a;\omega)|_{a=\mu_{\theta}(s)}]
\end{align*}
with $\int_{S}\sum_{t=0}^{\infty}\gamma^{t}p_{1}(s)p(s\rightarrow s',t,\mu_{\theta})ds=\rho^{\mu_{\theta}}(s')$\\

\end{document}